\documentclass[10pt,twocolumn,letterpaper]{article}

\usepackage[pagenumbers]{cvpr}

\usepackage{microtype}
\usepackage{makecell}

\makeatletter
\patchcmd{\paragraph}
  {3.25ex \@plus1ex \@minus.2ex}
  {2ex \@plus.4ex \@minus.2ex}
  {}{\PackageWarning{mvt}{Paragraph spacing patch failed}}
\makeatother

\usepackage{amsthm}
\newtheorem{theorem}{Theorem}
\newtheorem{proposition}[theorem]{Proposition}
\newtheorem{lemma}{Lemma}
\theoremstyle{definition}
\newtheorem{assumption}{Assumption}

\theoremstyle{plain}
\newtheorem{corollary}{Corollary}
\usepackage{tikz}
\usetikzlibrary{arrows.meta,positioning,calc,shadows}

\definecolor{viewtext}{HTML}{2864A0}
\definecolor{viewaudio}{HTML}{168578}
\definecolor{viewimage}{HTML}{B6692C}
\definecolor{viewpurple}{HTML}{8B3F92}
\usepackage{pgfplots}
\usepgfplotslibrary{groupplots}
\pgfplotsset{compat=1.18}

\usepackage{algorithm}
\usepackage{algpseudocode}

\newcommand{\bad}[1]{\fcolorbox{red!45!black!35}{red!9}{\strut #1}}
\newcommand{\good}[1]{\fcolorbox{viewaudio!70}{viewaudio!12}{\strut #1}}
\newcommand{\task}[2]{{\color{#1}\textbf{#2}}}

\newcommand{\one}{\mathbf 1}
\newcommand{\winit}{\omega_{\mathrm{init}}}
\newcommand{\Dis}{\mathrm{Dis}}

\newcommand{\QAGain}{+15.2}
\newcommand{\ForgetGain}{+40.7}
\newcommand{\EditGain}{+25.2}

\newcommand{\gainTMAvg}{+9.0}

\newcommand{\gainTWQA}{+15.2}

\newcommand{\gainTWFQ}{+40.7}

\newcommand{\gainTWGen}{+25.2}

\newcommand{\gainTWAvg}{+14.9}

\newcommand{\BackboneQAGain}{+17.8}

\newcommand{\BackboneForgetGain}{+4.3}

\newcommand{\BackboneEditGain}{+23.3}

\newcommand{\cmBaseFQImage}{2.1}
\newcommand{\cmMVTFQImage}{10.8}

\definecolor{cvprblue}{rgb}{0.21,0.49,0.74}
\usepackage[pagebackref,breaklinks,colorlinks,allcolors=cvprblue]{hyperref}

\def\confName{CVPR}
\def\confYear{2027}

\title{Multimodal View Training: Learning More\\ Through Different Modalities}

\author{%
Devang Kulshreshtha$^{1}$ \quad
Priyanka Kaswan$^{2}$\thanks{Work done independently of the authors' respective employers.} \quad
Minh Pham$^{3}$\footnotemark[1]\\
Chinmay Hegde$^{1}$ \quad Niv Cohen$^{4}$\\[4pt]
$^{1}$New York University \quad $^{2}$Snorkel AI \quad $^{3}$Meta\\
$^{4}$Technion -- Israel Institute of Technology\\
{\tt\small devangkulshreshtha12@gmail.com}
}

\begin{document}
\maketitle
\suppressfloats[t]
\begin{abstract}
Can a language model learn more from the same information simply by encountering it in different forms? We answer this in the affirmative by introducing \emph{Multimodal View Training (MVT)}, a simple training approach that presents identical information through text, speech, and images, without requiring new facts or supervision. Starting from a text sample, MVT generates corresponding audio and rendered-text image views, and trains a multimodal language model to produce the same target output across all three views. To further bring these benefits to text-only large language models (LLMs), we train lightweight connector modules to pretrained audio and vision encoders, allowing the LLM to learn from multimodal inputs while retaining its text-only architecture at inference. Across question answering, machine unlearning, and knowledge editing, adaptive MVT improves QA recall by \QAGain{} percentage points, forget quality by \ForgetGain{} points, and editing generality by \EditGain{} points over text-only training. Combining MVT with standard augmentation further improves QA and editing generality. Our theoretical analysis illustrates how training across modalities can shift weight updates toward shared factual content and reduce their reliance on features specific to each view.

\end{abstract}

\definecolor{ovTextFill}{HTML}{E5EFF9}  \definecolor{ovTextDraw}{HTML}{447BAA}  \definecolor{ovTextIcon}{HTML}{1A56B0}
\definecolor{ovAudioFill}{HTML}{FDEBDE} \definecolor{ovAudioDraw}{HTML}{E8802C} \definecolor{ovAudioIcon}{HTML}{E8802C}
\definecolor{ovPrintFill}{HTML}{F1EAF8} \definecolor{ovPrintDraw}{HTML}{9463B3} \definecolor{ovPrintIcon}{HTML}{8850A6}
\definecolor{ovBoxFill}{HTML}{EFEFEF}   \definecolor{ovBoxDraw}{HTML}{4D4D4D}
\definecolor{ovHeadFill}{HTML}{E5E5E5}  \definecolor{ovLine}{HTML}{333333}
\begin{figure}[t]
\centering
\resizebox{\columnwidth}{!}{%
\begin{tikzpicture}[
  font=\small,
  shadowed/.style={drop shadow={shadow xshift=0.6pt, shadow yshift=-0.8pt, opacity=0.18, fill=black}},
  gbox/.style={fill=ovBoxFill, draw=ovBoxDraw, line width=0.7pt, rounded corners=2.5pt, shadowed, align=center, inner sep=2pt},
  view/.style={draw=none, rounded corners=8pt, minimum width=4.5cm, minimum height=0.95cm},
  viewtag/.style={anchor=west, rounded corners=4pt, text=white, font=\sffamily\bfseries\scriptsize, inner xsep=4pt, inner ysep=2pt},
  arr/.style={draw=ovLine, line width=0.8pt, -{Latex[length=2.2mm,width=1.7mm]}},
  ln/.style={draw=ovLine, line width=0.8pt},
  icon/.style={line width=1.1pt, line join=round, line cap=round},
]
\node[gbox, minimum width=1.25cm, minimum height=1.0cm] (x) at (0.6,1.2) {Example\\$x$};

\node[view, fill=ovTextFill]  (t) at (4.15,2.4) {};
\node[view, fill=ovAudioFill] (a) at (4.15,1.2) {};
\node[view, fill=ovPrintFill] (i) at (4.15,0.0) {};

\node[viewtag, fill=ovTextDraw] at ($(t.north west)+(0.2,0)$) {TEXT};
\node[viewtag, fill=ovAudioDraw] at ($(a.north west)+(0.2,0)$) {AUDIO};
\node[viewtag, fill=ovPrintDraw] at ($(i.north west)+(0.2,0)$) {IMAGE};
\node[font=\sffamily\small, inner sep=0pt] at (4.15,2.32) {Where was Mira born?};

\begin{scope}[shift={(2.38,1.12)}, scale=0.011, yscale=-1, icon, draw=ovAudioIcon]
  \path[fill=ovAudioIcon, draw=none] (-29,-10) -- (-17,-10) -- (1,-24) -- (1,24) -- (-17,10) -- (-29,10) -- cycle;
  \draw (5,0) ++(-45:15) arc[start angle=-45, end angle=45, radius=15];
\end{scope}
\foreach \h [count=\j from 0] in {0.06,0.10,0.08,0.14,0.12,0.18,0.15,0.20,0.17,0.23,0.16,0.12,0.18,0.24,0.20,0.14,0.22,0.26,0.18,0.23,0.15,0.12,0.19,0.24,0.20,0.14,0.17,0.22,0.18,0.12,0.20,0.25,0.17,0.14,0.21,0.18,0.11,0.16,0.13,0.18,0.14,0.10,0.13,0.08,0.06}{
  \draw[draw=ovAudioIcon, line width=1.3pt, line cap=round]
    ({2.90+0.07*\j},{1.12-\h}) -- ({2.90+0.07*\j},{1.12+\h});
}
\node[fill=white, minimum width=3.95cm, minimum height=0.60cm,
      font=\rmfamily\small, inner sep=2pt] at (4.15,-0.07) {Where was Mira born?};

\node[gbox, fill=ovTextFill, draw=ovTextDraw, minimum width=3.3cm, minimum height=0.8cm] (te) at (8.85,2.4) {Text embeddings};
\node[gbox, fill=ovAudioFill, draw=ovAudioDraw, minimum width=1.25cm, minimum height=0.8cm] (ae) at (7.825,1.2) {Audio\\encoder};
\node[gbox, fill=ovAudioFill, draw=ovAudioDraw, minimum width=1.55cm, minimum height=0.8cm] (ap) at (9.875,1.2) {Connector};
\node[gbox, fill=ovPrintFill, draw=ovPrintDraw, minimum width=1.25cm, minimum height=0.8cm] (ve) at (7.825,0.0) {Vision\\encoder};
\node[gbox, fill=ovPrintFill, draw=ovPrintDraw, minimum width=1.55cm, minimum height=0.8cm] (vp) at (9.875,0.0) {Connector};

\node[gbox, minimum width=1.7cm, minimum height=1.9cm, font=\normalsize\bfseries] (f) at (12.7,1.2)
  {Shared\\model\\[2pt]{\normalfont$f_\omega$}};

\draw[arr] (x.east) -- (a.west);
\draw[arr] ($(x.east)+(0.2,0)$) |- (t.west);
\draw[arr] ($(x.east)+(0.2,0)$) |- (i.west);
\draw[arr] (t.east) -- (te.west);
\draw[arr] (a.east) -- (ae.west);  \draw[arr] (ae.east) -- (ap.west);
\draw[arr] (i.east) -- (ve.west);  \draw[arr] (ve.east) -- (vp.west);
\draw[ln, rounded corners=3pt] (te.east) -| (11.0,1.2);
\draw[ln, rounded corners=3pt] (vp.east) -| (11.0,1.2);
\draw[arr] (ap.east) -- (f.west);

\node[gbox, minimum width=2.3cm, minimum height=0.9cm] (y) at (1.55,-1.85) {\textbf{Supervision} $y$\\\textit{``Luma''}};

\node[draw=ovBoxDraw, fill=white, line width=0.8pt, rounded corners=4pt, shadowed,
      minimum width=9.7cm, minimum height=1.55cm] (obj) at (8.7,-1.85) {};
\node[fill=ovHeadFill, rounded corners=3pt, minimum width=9.4cm, minimum height=0.38cm,
      font=\small\bfseries, inner sep=0pt] at (8.7,-1.32) {Same task objective};
\node[fill=ovTextFill,  rounded corners=3pt, minimum width=2.9cm, minimum height=0.62cm] at (5.6,-2.05)
  {$\mathcal{L}\bigl(f_\omega(z_T),y\bigr)$};
\node[fill=ovAudioFill, rounded corners=3pt, minimum width=2.9cm, minimum height=0.62cm] at (8.7,-2.05)
  {$\mathcal{L}\bigl(f_\omega(z_A),y\bigr)$};
\node[fill=ovPrintFill, rounded corners=3pt, minimum width=2.9cm, minimum height=0.62cm] at (11.8,-2.05)
  {$\mathcal{L}\bigl(f_\omega(z_I),y\bigr)$};

\draw[arr] (y.east) -- (obj.west);
\draw[arr] (12.3,0.25) -- (12.3,-1.075);
\draw[arr, dashed] (13.1,-1.075) -- node[right, font=\footnotesize\itshape, align=left, inner sep=2pt] {shared\\update} (13.1,0.25);
\end{tikzpicture}}
\caption{\textbf{MVT: same supervision, different views.} Text,
speech, and printed-text views share target $y$. Their losses are
combined before one LoRA update to the shared LLM.
Connectors are pretrained or learned via text reconstruction,
then frozen during MVT.}
\label{fig:overview}
\end{figure}
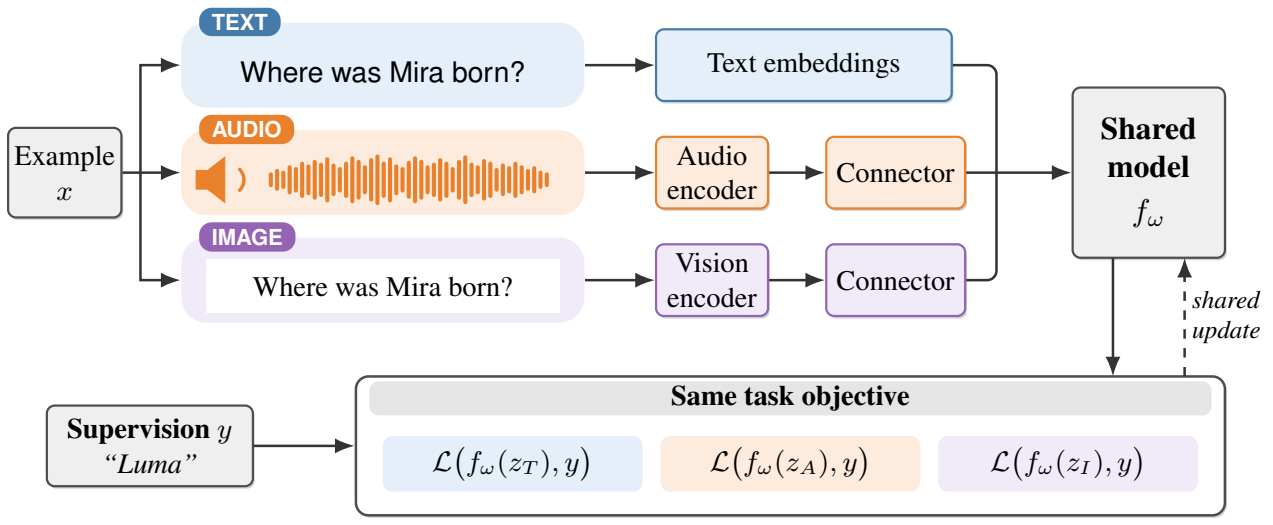
\section{Introduction}
\label{sec:intro}

Fine-tuning allows large language models (LLMs) to adapt to new tasks using relatively few training examples, making it important to learn effectively from each example~\cite{jain2024neftune}. However, learning from an example should go beyond memorizing how it was presented. For instance, a model trained to answer a question should ideally answer it correctly even when the question is worded differently. Data augmentation improves this generalization by exposing the model to different versions of same training example~\cite{najafi2024riff}.

Existing augmentation methods introduce variation by changing the wording of an example or perturbing its embedding representation. Text-based methods, such as paraphrasing, word substitution, and back-translation~\cite{wei2019eda,xie2019uda,najafi2024riff}, generate alternative expressions of the same content. However, these transformations may unintentionally change the meaning or omit important details. Embedding-based perturbation methods, such as MixText, NEFTune, and SymNoise~\cite{chen2020mixtext,jain2024neftune,yadav2025symnoise} modify internal representations, typically introducing only small and local variations. This motivates us to explore a different form of augmentation that can provide substantially different representations while preserving the original content.

Multimodal learning has demonstrated the value of combining information from text, speech, and images~\cite{ngiam2011multimodal,liu2023llava,tang2024salmonn}, but typically requires collecting or aligning observations from multiple modalities. A simpler alternative is to generate multiple views directly from text: a question can be converted into speech or printed as an image while preserving its original content. These views contain the same information yet produce different representations inside the model. This raises a natural question: \textit{can such diverse views improve an LLM's performance on text-based tasks without additional task information?}

We answer this question with \emph{Multimodal View Training} (MVT). For
each text training example, MVT synthesizes it as speech and renders its
text as a printed image, while keeping the original target output for all three views
(Figure~\ref{fig:overview}). We first propose MVT using a
multimodal language model (MLM) whose text, audio, and image pathways feed
into a shared LLM. We train only the shared LLM using the average task
loss across the three views, while keeping the modality-specific
components frozen. After training, we retain only the LLM and its
text input pathway, resulting in a text-only model with no additional
inference cost.

To extend MVT to text-only LLMs that lack multimodal input pathways,
we temporarily equip them with pretrained audio and vision encoders. We then add lightweight MLP connectors to project the different modalities to the same space ~\cite{li2023blip2,liu2023llava,tang2024salmonn}. 
We first train the connectors to align
the modality representations with the frozen LLM, and then apply MVT
to update the LLM while keeping these auxiliary components fixed.
The auxiliary components are discarded after training, so the final
model remains text-only.

We evaluate MVT across question answering, machine unlearning, and knowledge editing. Compared with text-only training, adaptive MVT improves QA recall by \QAGain{} percentage points, forget quality by \ForgetGain{} points, and editing generality by \EditGain{} points. Adding MVT to each text augmentation baseline also improves QA and editing generality (Section~\ref{sec:results}). We also provide a theoretical analysis explaining
the advantage of using MVT (Section~\ref{sec:theory}). Together, these results demonstrate that multiple modalities can serve as effective training views, improving text-based learning without requiring multimodal inputs at inference time.

\section{Multimodal View Training}
\label{sec:method}

MVT uses different modality presentations of the same text input
to train a shared language model. We first define the modality
pathways and training objective, then describe how to extend MVT
to text-only LLMs.

\subsection{Problem Formulation}
\label{sec:setup}

Let $D=\{(x_i,y_i)\}_{i=1}^N$ be a dataset of text inputs $x_i$
and their task-specific supervision $y_i$, and let $\ell$ denote
the task loss. We seek to improve a shared language model by
presenting each input through three modality pathways while
keeping the supervision unchanged. We index the modalities by
$m\in\mathcal M=\{T,A,I\}$ for text, audio, and printed-text
images, respectively.

\paragraph{Rendering an input.}
A renderer $r_m$ produces the presentation for modality $m$.
For a text input $x$, $r_T(x)=x$ retains the text, $r_A(x)$
synthesizes its speech, and $r_I(x)$ renders its words as a
printed-text image. All views preserve the original
linguistic content and use the same task supervision.

\paragraph{Encoding a view.}
Each rendered view is converted into an embedding sequence
that the LLM can process. For speech and images, a pretrained
modality encoder $E_m$ extracts features, and a connector $\psi_m$
maps them into the LLM's input embedding space. We denote this
combined operation by $\phi_m$, giving
\begin{equation}
z_m(x)=\phi_m(r_m(x))=
\begin{cases}
\operatorname{Embed}(x), & m=T,\\
\psi_m(E_m(r_m(x))), & m\in\{A,I\},
\end{cases}
\label{eq:pathway}
\end{equation}
where $\operatorname{Embed}$ denotes the LLM's tokenizer and
token embeddings. We refer to the complete route from $x$ to $z_m(x)$
as a \emph{modality pathway}. In  multimodal models
such as Qwen3-Omni, the encoders and connectors are already
aligned with the LLM~\cite{qwen3omni}, so they can be used
directly for MVT.

\paragraph{Shared language model.}
Let $f_\omega$ denote the pretrained LLM augmented with
trainable LoRA parameters $\omega$ used for the fine-tuning task (e.g., Question Answering)~\cite{hu2022lora}.
During MVT, we update only $\omega$ while keeping all
other components frozen.

\subsection{Training Objective}
\label{sec:multiview}

Each modality pathway produces a different representation of
the same input. We compute the original task loss through
each pathway and use the resulting gradients to update the
shared LLM. For a minibatch $B$, the loss for modality $m$ is
\begin{equation}
L_m(\omega;B)=\frac{1}{|B|}\sum_{(x,y)\in B}
\ell\big(f_\omega(z_m(x)),y\big).
\label{eq:viewloss}
\end{equation}
The task loss and supervision are identical across views;
only the input representation changes.

\paragraph{Combining the views.}
At each training step, we pass the same minibatch through
all three pathways. We combine their losses using
\begin{equation}
\mathcal{L}_{\mathrm{MVT}}(\omega;B)
=\tau\log\left(
\frac{1}{3}\sum_{m\in\mathcal M}
\exp\left(\frac{L_m(\omega;B)}{\tau}\right)
\right).
\label{eq:mvtobjective}
\end{equation}
For $\tau\to\infty$, this reduces to the uniform average
of the three losses, giving equal importance to each view.
However, different modality pathways may have different
task losses. We therefore also consider adaptive weighting,
where a finite temperature $\tau>0$ assigns greater weight
to views having higher-loss. Smaller values of $\tau$ concentrate
more weight on the highest-loss view. The resulting gradient
is a weighted combination of the view gradients, with
weights given by $\mathbf w=\operatorname{softmax}(\mathbf L/\tau)$,
where $\mathbf L=(L_T,L_A,L_I)$. This objective is motivated
by group-level tilted loss~\cite{li2021tilted}.

\paragraph{Inference.}
\label{sec:augmentation-inference}
At inference, a text query uses only the text embeddings and
LoRA-adapted LLM. The audio and image components are no longer
needed and can be discarded, so MVT adds no inference-time
computation.

\subsection{Modality Alignment via Text Reconstruction}
\label{sec:anybackbone}

MVT applies directly to pretrained multimodal LMs with
existing audio and image pathways. To extend it to a
text-only LLM, we attach pretrained speech and image encoders
and learn lightweight connectors that make their outputs
interpretable by the LLM.

Following BLIP-2~\cite{li2023blip2}, we use the connector
$\psi_m$ with trainable parameters for each
auxiliary modality, audio and images, $m\in\{A,I\}$. We train these connectors
to reconstruct the original text from the encoded speech
or image, while keeping the encoders and LLM frozen.

For training the alignment connector, we use a set of unsupervised sentences $s \in D_{\mathrm{align}}$,
disjoint from the downstream task data. Similar to before, each sentence $s$ will go through full modality pathway to produce $z_m(s)$. Using
$s$ as the target as well, we minimize
\begin{equation}
\mathcal{L}_{\mathrm{align},m}(\psi_m)
=\mathbb{E}_{s\sim D_{\mathrm{align}}}
\ell_{\mathrm{CE}}\big(f_\omega(z_m(s)),s\big),
\label{eq:align}
\end{equation}
where the cross-entropy is computed over tokens in $s$, meaning the model is asked to reconstruct the same text via audio/image input. Only the connector parameters $\psi_m$
are updated during this stage.

After alignment, we freeze the encoders and connectors and train the LLM’s LoRA parameters using MVT. We then discard the auxiliary components and retain the text-only LLM. Algorithm~\ref{alg:mvt} summarizes both stages.

\begin{algorithm}[t]
\caption{Multimodal View Training (MVT)}
\label{alg:mvt}
\begin{algorithmic}[1]
\Require LoRA LLM $f_\omega$;
task data $D$; renderers $\{r_m\}$;
auxiliary encoders/connectors $\{E_m,\psi_m\}_{m\in\{A,I\}}$; optimizer $\mathcal O$;
training steps $T$; temperature schedule $\{\tau_t\}$;
optional alignment data $D_{\mathrm{align}}$
\Statex \textbf{Stage 1: Optional connector alignment}
\State Set $\phi_T\gets\operatorname{Embed}$
\State Build $\phi_m$ from $E_m$ then $\psi_m$, $m\in\{A,I\}$
\For{each $m\in\{A,I\}$ with an unaligned connector}
    \State Train only $\psi_m$ using Eq.~\eqref{eq:align}
\EndFor
\Statex \textbf{Stage 2: LoRA-only task training}
\State Freeze all pathway parameters and original LLM weights
\For{$t=1,\ldots,T$}
    \State Sample minibatch $B\sim D$ for all views
    \State Compute $L_m(\omega;B)$ for all $m$
           using Eq.~\eqref{eq:viewloss}
    \State Compute $\mathcal L$ using Eq.~\eqref{eq:mvtobjective}
           with $\tau=\tau_t$
    \State $d\gets\nabla_\omega\mathcal L$
    \State $\omega\gets\mathcal O(\omega,d)$
\EndFor
\State \Return Text LLM with trained LoRA $\omega$
\end{algorithmic}
\end{algorithm}

\section{Why can alternative views help a text model?}\label{sec:theory}

When the model learns that the answer to ``Where was Mira born?'' is ``Luma''
(Figure~\ref{fig:overview}), the weight update can strengthen two kinds of directions. The
\emph{content direction} is tied to the fact and can make the correct
answer more likely across views. The \emph{surface directions} reflect
cues tied to the training question's exact wording or modality. When the
question is reworded as ``What is Mira's birthplace?'', the content direction
still helps the model answer correctly, while the original text's surface
direction may help less. We present the main results below and give the full
proofs in Appendix~\ref{sec:appendix-theory} of the supplementary material.

\subsection{Setup and Assumptions}

Below, we first show when training on multiple views shifts the optimizer's
update toward shared content and improves transfer to paraphrased questions
(Theorem~\ref{thm:diversity}, Section~\ref{sec:theory-transfer}). We then show
that the MVT loss splits into the cross-entropy of the combined prediction
and a term that encourages agreement across views
(Proposition~\ref{prop:agreement}, Section~\ref{sec:theory-agreement}).

To set up the first result, we measure learning by how much fine-tuning
raises the log probability of the correct answer. For a question $x$ with answer $y=(y_1,\dots,y_L)$,
let $z_m$ be the input representation for view $m$ (Section~\ref{sec:setup})
and $\omega$ the LoRA parameters we aim to fine-tune. We call the log probability of the
full answer the \emph{answer score}:\vspace{-0.5em}
\begin{equation}\label{eq:answer-score}
h(z_m;\omega)=\sum_{t=1}^{L}\log p_\omega(y_t\mid z_m,y_{<t}).\vspace{-0.5em}
\end{equation}
A higher
answer score means the correct answer is more likely. Its
negative is the total cross-entropy loss.

To see how a weight update affects the answer score, we evaluate the score's
gradient for each view at initialization. We hold these initial gradients fixed,
motivated by evidence that gradient features can remain approximately stable in some
fine-tuning settings~\citep{malladi2023kernel}.

Let $\winit$ be the initial LoRA parameters and $\delta$ the weight update
since initialization, so $\omega=\winit+\delta$.
For a view with input representation $z$, let $u(z)$ denote
the gradient of the answer score with respect to the LoRA parameters,
evaluated at initialization. We call $u(z)$ the \emph{gradient feature}.
\begin{equation}\label{eq:gradient-feature}
u(z)=\left.\nabla_\omega h(z;\omega)\right|_{\omega=\winit}.
\end{equation}
A first-order Taylor approximation around $\winit$ gives
\begin{equation}\label{eq:lin}
h(z;\winit+\delta)-h(z;\winit)\approx u(z)^\top\delta.
\end{equation}
The inner product $u(z)^\top\delta$ predicts how much the weight update
raises or lowers the answer score. The approximation is local because it omits
higher-order terms, which can matter as the
weights move away from initialization.

We refer to each training question-answer pair $(x_i,y_i)$ as \emph{fact} $i$.
When we present question $x_i$ in view $m$, the model receives the input
representation $z_{i,m}=\phi_m(r_m(x_i))$ from Eq.~\eqref{eq:pathway}.  Eq.~\eqref{eq:gradient-feature} at $z_{i,m}$ gives the view gradient
feature $u_{i,m}$.

\begin{assumption}[Shared and view-specific gradient components]\label{ass:S}
Each view presents the same fact but may offer different cues for answering
the question. We model the shared part of the gradient feature by the
\emph{content direction} $a_i$, and the part tied to the wording or modality
of view $m$ by the \emph{surface direction} $e_{i,m}$. Together they give
$u_{i,m}=a_i+e_{i,m}$. Surface directions for the same fact can overlap
across views. Their dot products $(G_i)_{mn}=e_{i,m}^{\top}e_{i,n}$ form
the Gram matrix $G_i$. Appendix~\ref{sec:appendix-theory} gives additional assumptions.
\end{assumption}

\subsection{How adding views affects transfer}\label{sec:theory-transfer}

MVT trains the model to perform well on multiple views of each example.
We show how adding training views improves generalization to paraphrased
questions. To fulfill this objective, we aim to first increase the answer score by
$\gamma_i>0$ on every training view of fact $i$.\footnote{Appendix~\ref{sec:appendix-theory}
also considers increase in score $\gamma_{i,m}$ varying by view $m$.}
Let $\delta^\star$ denote the resulting weight update from gradient descent,
and let $\alpha^\star=a_i^\top\delta^\star$ measure how much of it lies along
the shared content direction. 

To assess generalization, we use Eq.~\eqref{eq:lin} to estimate how much
the same update raises the answer score on a paraphrase of the original
text question. The \emph{transfer ratio} $\nu_i$ is the answer score increase
on the paraphrased question divided by the training gain $\gamma_i$.
We assume that the paraphrase's gradient feature keeps the content
direction $a_i$ in full because the question still asks for the same fact.
Paraphrasing however can change surface cues, so we assume that only a fraction
$\rho_i$ of the original text's surface direction $e_{i,T}$ carries over. 

Let $\Pi_i=\one^\top G_i^{-1}\one$ be an intermediate quantity, where $G_i$ is the gram matrix defined under Assumption~\ref{ass:S}.
Theorem~\ref{thm:diversity} first shows that adding a view to the training can raise $\Pi_i$. Then it further shows that as $\Pi_i$ rises, both the shared content $\alpha^\star$ and the transfer ratio $\nu_i$ increase.

\begin{theorem}[Transfer to paraphrased questions]\label{thm:diversity}
Under Assumption~\ref{ass:S}, the content term $\alpha^\star$ and
transfer ratio $\nu_i$ satisfy
\begin{align}
\alpha^\star &= \frac{\gamma_i\Pi_i}{1+\Pi_i},\label{eq:content-component}\\
\nu_i &= 1-\frac{1-\rho_i}{1+\Pi_i}.\label{eq:shortfall}
\end{align}
If the views used by MVT for fact $i$ are extended by one, the updated
cost $\Pi_i'$ satisfies $\Pi_i'\geq\Pi_i$.
\end{theorem}

Appendix~\ref{sec:appendix-theory} gives the detailed proof. For intuition, consider an added view whose surface direction
is orthogonal to the existing ones. Raising its answer score using
surface cues alone requires a larger weight update. The shared content
direction helps every view; therefore, more of the update goes toward shared content, as
Eq.~\eqref{eq:content-component} shows. The effect of that shared content
carries over in full to a paraphrased question, which explains why
transfer can improve in Eq.~\eqref{eq:shortfall}.

\subsection{How MVT encourages agreement across views}\label{sec:theory-agreement}

Different views of the same question can predict different token
probabilities. To see how the MVT loss captures this disagreement,
consider a single target answer token $y$. For view $m$, let $s_m$ be the
logits (the scores before softmax) and $p_m=\mathrm{softmax}(s_m)$ the
predicted token probabilities. We take the weighted average of the logits,
$\bar s=\sum_{m\in\mathcal M}w_m s_m$, and apply softmax to obtain the
combined prediction $\bar p=\mathrm{softmax}(\bar s)$. Proposition~\ref{prop:agreement} shows that the MVT loss can be decomposed into a loss on the combined
prediction and a term that measures disagreement across views (full proof in Appendix~\ref{sec:appendix-theory}).

\begin{proposition}[Cross-entropy decomposition across views]\label{prop:agreement}
\begin{equation}\label{eq:agreement}
\begin{split}
\sum_{m\in\mathcal M} w_m\,\ell_{\mathrm{CE}}(s_m,y)
={}&\ell_{\mathrm{CE}}(\bar s,y)\\
&+\underbrace{\sum_{m\in\mathcal M} w_m\,D_{\mathrm{KL}}(\bar p\,\|\,p_m)}_{\text{disagreement}}.
\end{split}
\end{equation}
\end{proposition}
By penalizing the disagreement in Eq.~\eqref{eq:agreement}, the MVT loss
encourages updates to the shared LoRA parameters that bring the views'
predicted token distributions closer together. This identity is the
likelihood form of the ensemble ambiguity
decomposition~\citep{krogh1995neural,heskes1998bias,wood2023unified}; 
we restate it for MVT.
\section{Experimental Setup}
\label{sec:experiments}

\subsection{Datasets and Metrics}
\label{sec:data}\label{sec:metrics}
We evaluate MVT on three textual tasks: question answering,
machine unlearning, and knowledge editing. Each task uses its
original training objective and evaluation metrics.

\begin{table*}[t]
\centering\footnotesize
\setlength{\tabcolsep}{2pt}
\begin{tabular*}{\textwidth}{@{\extracolsep{\fill}}lccccccc@{}}
\toprule
& QA $\uparrow$ & \multicolumn{2}{c}{Unlearning $\uparrow$} & \multicolumn{3}{c}{Editing $\uparrow$} & \\
\cmidrule(lr){2-2}\cmidrule(lr){3-4}\cmidrule(lr){5-7}
Method & Recall & Forget quality & Model utility & Reliability & Generality & Locality & Average $\uparrow$ \\
\midrule
Text-only & $42.3\!\pm\!0.6$ & $13.8\!\pm\!4.2$ & $61.5\!\pm\!1.3$ & $95.2\!\pm\!0.5$ & $61.8\!\pm\!0.1$ & $17.5\!\pm\!0.2$ & $48.7$ \\
\quad + MVT & $50.7\!\pm\!0.7$ & $27.1\!\pm\!9.5$ & $\mathbf{65.9}\!\pm\!0.4$ & $\mathbf{99.7}\!\pm\!0.0$ & $\mathbf{87.2}\!\pm\!0.8$ & $15.8\!\pm\!1.1$ & $57.7$ \\
\quad + MVT (adaptive) & $\mathbf{57.5}\!\pm\!0.7$ & $\mathbf{54.5}\!\pm\!20.2$ & $64.8\!\pm\!0.5$ & $\mathbf{99.7}\!\pm\!0.0$ & $87.0\!\pm\!2.0$ & $\mathbf{18.1}\!\pm\!1.9$ & $\mathbf{63.6}$ \\
\midrule
Paraphrasing & $54.6\!\pm\!1.6$ & $13.8\!\pm\!0.4$ & $65.1\!\pm\!0.5$ & $\mathbf{99.7}\!\pm\!0.1$ & $86.7\!\pm\!0.8$ & $\mathbf{17.4}\!\pm\!2.2$ & $56.2$ \\
\quad + MVT & $67.0\!\pm\!1.0$ & $\mathbf{14.2}\!\pm\!4.6$ & $\mathbf{67.8}\!\pm\!0.7$ & $\mathbf{99.7}\!\pm\!0.1$ & $93.2\!\pm\!1.0$ & $14.4\!\pm\!1.9$ & $59.4$ \\
\quad + MVT (adaptive) & $\mathbf{73.9}\!\pm\!1.1$ & $8.8\!\pm\!2.8$ & $67.0\!\pm\!0.4$ & $\mathbf{99.7}\!\pm\!0.0$ & $\mathbf{96.3}\!\pm\!0.2$ & $14.2\!\pm\!1.1$ & $\mathbf{60.0}$ \\
\midrule
NEFTune & $42.4\!\pm\!1.4$ & $20.2\!\pm\!9.2$ & $60.4\!\pm\!1.6$ & $92.7\!\pm\!0.7$ & $62.4\!\pm\!1.0$ & $\mathbf{16.4}\!\pm\!0.8$ & $49.1$ \\
\quad + MVT & $52.8\!\pm\!0.8$ & $39.4\!\pm\!4.2$ & $\mathbf{65.6}\!\pm\!0.8$ & $99.0\!\pm\!0.2$ & $86.8\!\pm\!1.3$ & $15.8\!\pm\!0.3$ & $59.9$ \\
\quad + MVT (adaptive) & $\mathbf{55.4}\!\pm\!0.9$ & $\mathbf{46.6}\!\pm\!26.1$ & $65.0\!\pm\!0.9$ & $\mathbf{99.3}\!\pm\!0.1$ & $\mathbf{88.8}\!\pm\!3.1$ & $14.6\!\pm\!0.9$ & $\mathbf{61.6}$ \\
\midrule
SymNoise & $43.8\!\pm\!1.1$ & $25.8\!\pm\!4.5$ & $59.7\!\pm\!0.6$ & $96.9\!\pm\!0.0$ & $68.1\!\pm\!1.7$ & $\mathbf{18.7}\!\pm\!0.9$ & $52.2$ \\
\quad + MVT & $48.7\!\pm\!0.4$ & $54.5\!\pm\!4.6$ & $\mathbf{65.3}\!\pm\!0.4$ & $98.6\!\pm\!0.3$ & $86.1\!\pm\!2.2$ & $16.7\!\pm\!0.7$ & $61.7$ \\
\quad + MVT (adaptive) & $\mathbf{54.5}\!\pm\!0.5$ & $\mathbf{71.3}\!\pm\!4.7$ & $64.3\!\pm\!0.6$ & $\mathbf{99.7}\!\pm\!0.0$ & $\mathbf{90.0}\!\pm\!1.7$ & $16.7\!\pm\!2.0$ & $\mathbf{66.1}$ \\
\midrule
\textit{Average of the four} & $45.8$ & $18.4$ & $61.7$ & $96.1$ & $69.8$ & $\mathbf{17.5}$ & $51.6$ \\
\quad + MVT & $54.8$ & $33.8$ & $\mathbf{66.2}$ & $99.3$ & $88.3$ & $15.7$ & $59.7$ \\
\quad + MVT (adaptive) & $\mathbf{60.3}$ & $\mathbf{45.3}$ & $65.3$ & $\mathbf{99.6}$ & $\mathbf{90.5}$ & $15.9$ & $\mathbf{62.8}$ \\
\bottomrule
\end{tabular*}

\caption{\textbf{MVT improves learning, unlearning, and editing.}
Each baseline is compared with its MVT-augmented version.
Scores are percentages; bold marks the best mean in each three-row group.
The final three rows report averages of the four method means; the last column
averages the six metric means.}
\label{tab:main}\label{tab:qa}\label{tab:unlearn}\label{tab:edit}
\end{table*}

\begin{table*}[t]
\centering\footnotesize
\setlength{\tabcolsep}{4pt}
\renewcommand{\arraystretch}{1.1}
\setlength{\fboxsep}{1.2pt}
\begin{tabular}{@{}>{\raggedright\arraybackslash}p{.095\textwidth}
>{\raggedright\arraybackslash}p{.30\textwidth}
>{\raggedright\arraybackslash}p{.335\textwidth}
>{\raggedright\arraybackslash}p{.20\textwidth}@{}}
\toprule
& \task{viewtext}{Unlearning} (TOFU) & \task{viewimage}{QA} (TOFU; father / mother) & \task{viewpurple}{Editing} (ZsRE) \\
\textit{Query} & Are Aysha Al-Hashim's books available in languages other than English? &
What are the occupations of Hina Ameen's mother and father? &
Which planet is Ulysses Colles located? \\
\textit{Target} & forget: Arabic & Real Estate Agent / Doctor & Mars $\to$ Moon \\
\midrule
Paraphrasing & \dots originally in \bad{Arabic}\dots &
\bad{hairdresser} / \bad{lawyer} &
\bad{Mars} \\
NEFTune & \dots including \bad{Arabic}\dots &
\bad{translator} / \bad{midwife} &
\bad{Pluto} \\
SymNoise & \dots including \bad{Arabic}\dots &
\bad{Veterinarian} / \bad{Research Scientist} &
\bad{Mercury} \\
\textbf{MVT (ours)} & \dots translated into \good{multiple languages}\dots &
\good{Real Estate Agent} / \good{Doctor} &
\good{Moon} \\
\bottomrule
\end{tabular}
\caption{\textbf{Qualitative comparison of training methods on QA, unlearning, and editing.} \bad{Red}: wrong answer; \good{green}: correct answer.}
\label{tab:qualitative}
\end{table*}

\paragraph{Question answering.}
We fine-tune on TOFU~\cite{eldan2023tofu} question--answer pairs
using cross-entropy and evaluate on held-out paraphrases of
the original questions. This tests whether the model can recall
learned facts when questions are phrased differently. We report
ROUGE-L recall~\cite{lin2004rouge} between generated answers
and reference answers.

\paragraph{Machine unlearning.}
Machine unlearning aims to suppress designated facts while
preserving general knowledge. We use TOFU~\cite{eldan2023tofu},
which contains question--answer pairs about fictitious authors.
We first fine-tune the model on all author facts. Starting from the same fine-tuned checkpoint, all methods unlearn the \texttt{forget05} split using SimNPO~\cite{fan2024simnpo}.
Following TOFU, we report:
\begin{itemize}
\item \textbf{Forget quality (FQ):} measures how closely the
unlearned model's behavior on forget questions matches that
of a reference model trained only on retained data.
Higher values indicate better forgetting.
\item \textbf{Model utility (MU):} measures retained performance
on the retain set, real authors, and world facts. It is the
harmonic mean of nine scores obtained from answer probability,
ROUGE-L recall, and correct-answer preference across these
three sets.
\end{itemize}

\paragraph{Knowledge editing.}
Knowledge editing aims to replace specific facts while preserving
unrelated knowledge. We use the MEND version of
ZsRE~\cite{levy2017zsre,mitchell2022fast} and fine-tune with
cross-entropy on source questions paired with edited
answers~\cite{zhu2020modifying}. We report three ROUGE-L
recall scores~\cite{meng2022rome,mitchell2022fast}:
\emph{reliability}, which measures whether the model produces
the edited answer on the original question;
\emph{generality}, which evaluates the same edit on
paraphrased questions; and \emph{locality}, which measures
performance on unrelated questions using their original
reference answers.

All evaluations use greedy decoding.
Appendices~\ref{sec:appendix-data} and ~\ref{sec:appendix-eval}
provide further dataset and metric details.
\subsection{Models and Training}
\label{sec:models}

\paragraph{Multimodal views.}
We generate speech using Microsoft Edge neural
TTS\footnote{\url{https://github.com/rany2/edge-tts}}
at 16\,kHz. Printed-text images are rendered using Pillow
with DejaVu Sans font, black text, and a white
$896\times448$ canvas.

\paragraph{Models.}
Our main model is Qwen3-Omni-30B-A3B-Instruct~\cite{qwen3omni}.
We fine-tune LoRA adapters~\cite{hu2022lora} on the attention
projections of its LLM, everything else kept frozen.

To evaluate MVT on text-only LLMs, we use Llama-3.1-8B
~\cite{grattafiori2024llama3}, Yi-1.5-9B~\cite{young2024yi},
OLMo-2-7B~\cite{olmo2024olmo2}, and
Qwen2.5-32B~\cite{qwen2025qwen25}. We equip each model
with a pretrained speech encoder (AuT, a 650M
Audio Transformer~\cite{qwen3omni}) and a pretrained image
encoder (SigLIP2-So400m, a 540M
ViT~\cite{tschannen2025siglip2}). Each encoder is connected
to the LLM through a lightweight connector aligned using
49k WikiText-103 sentences, following
Section~\ref{sec:anybackbone}. After alignment, we apply
the same MVT training procedure as for Qwen3-Omni.

All methods are trained to convergence, with checkpoints
selected using a shared early-stopping rule on held-out
paraphrased questions. Under this protocol, the
optimizer steps and training costs may differ across methods. Appendix~\ref{sec:appendix-training}
provides the training protocols and configs.

\subsection{Training Methods}
\label{sec:baselines}

Within each task, all methods use the same task loss and
target answers. We compare the following training methods:

\begin{itemize}
\item \textbf{Text-only:} LoRA training on the original
text questions.

\item \textbf{Paraphrasing:} trains on the original questions
and one DIPPER-generated paraphrase~\cite{krishna2023dipper}
per question, using the same target answer.

\item \textbf{NEFTune}~\cite{jain2024neftune}\textbf{:}
adds uniform noise to the input embeddings during training.

\item \textbf{SymNoise}~\cite{yadav2025symnoise}\textbf{:}
applies symmetric $\pm$ noise to the input embeddings
and averages the two losses.

\item \textbf{MVT:} trains on text, speech, and printed-text
image views with equal weighting
(Eq.~\eqref{eq:mvtobjective}, $\tau\to\infty$).

\item \textbf{MVT (adaptive):} uses temperature-controlled
weighting to emphasize views with higher task losses
(Eq.~\eqref{eq:mvtobjective}).

\item \textbf{Text augmentation + MVT:} combines MVT with
paraphrasing, NEFTune, or SymNoise, applying the additional
augmentation only to the text pathway.
\end{itemize}
We repeat each task-training configuration three times and report
mean $\pm$ standard deviation.
Appendix~\ref{sec:appendix-baselines} gives hyperparameters for the above methods.

\section{Results}
\label{sec:results}

\begin{table*}[t]
\centering\footnotesize
\setlength{\tabcolsep}{1.5pt}
\begin{tabular*}{\textwidth}{@{\extracolsep{\fill}}llcccccccccc@{}}
\toprule
& & \multicolumn{2}{c}{Speech alignment $\downarrow$} & \multicolumn{2}{c}{Image alignment $\downarrow$} & QA $\uparrow$ & \multicolumn{2}{c}{Unlearning $\uparrow$} & \multicolumn{3}{c}{Editing $\uparrow$} \\
\cmidrule(lr){3-4}\cmidrule(lr){5-6}\cmidrule(lr){7-7}\cmidrule(lr){8-9}\cmidrule(lr){10-12}
Model & Training & NLL & CER (\%) & NLL & CER (\%) & Recall & FQ & MU & Rel. & Gen. & Loc. \\
\midrule
Qwen3-Omni & Text-only & \multicolumn{4}{c}{\textit{native encoders}} & $42.3\!\pm\!0.6$ & $13.8\!\pm\!4.2$ & $61.5\!\pm\!1.3$ & $95.2\!\pm\!0.5$ & $61.8\!\pm\!0.1$ & $17.5\!\pm\!0.2$ \\
 & + MVT (adaptive) & \multicolumn{4}{c}{} & $\mathbf{57.5}\!\pm\!0.7$ & $\mathbf{54.5}\!\pm\!20.2$ & $\mathbf{64.8}\!\pm\!0.5$ & $\mathbf{99.7}\!\pm\!0.0$ & $\mathbf{87.0}\!\pm\!2.0$ & $\mathbf{18.1}\!\pm\!1.9$ \\
\midrule
Llama-3.1-8B & Text-only & 0.31 & 9.2 & 0.08 & 2.9 & $42.3\!\pm\!0.4$ & $14.2\!\pm\!9.5$ & $59.7\!\pm\!0.7$ & $98.1\!\pm\!0.2$ & $75.5\!\pm\!0.6$ & $\mathbf{18.6}\!\pm\!0.8$ \\
 & + MVT (adaptive) &  &  &  &  & $\mathbf{60.7}\!\pm\!0.8$ & $\mathbf{17.8}\!\pm\!7.4$ & $\mathbf{62.9}\!\pm\!0.4$ & $\mathbf{99.4}\!\pm\!0.2$ & $\mathbf{89.8}\!\pm\!0.3$ & $14.6\!\pm\!1.8$ \\
\midrule
Yi-1.5-9B & Text-only & 0.31 & 11.5 & 0.08 & 6.5 & $40.4\!\pm\!1.5$ & $32.8\!\pm\!27.6$ & $60.1\!\pm\!2.0$ & $97.4\!\pm\!0.3$ & $64.6\!\pm\!3.5$ & $13.4\!\pm\!4.1$ \\
 & + MVT (adaptive) &  &  &  &  & $\mathbf{61.1}\!\pm\!0.8$ & $32.8\!\pm\!16.9$ & $\mathbf{62.2}\!\pm\!0.7$ & $\mathbf{99.7}\!\pm\!0.1$ & $\mathbf{91.3}\!\pm\!0.9$ & $\mathbf{14.4}\!\pm\!1.2$ \\
\midrule
OLMo-2-7B & Text-only & 0.44 & 12.8 & 0.09 & 2.9 & $38.8\!\pm\!0.6$ & $13.6\!\pm\!4.5$ & $64.5\!\pm\!0.7$ & $98.9\!\pm\!0.1$ & $62.1\!\pm\!0.7$ & $\mathbf{8.6}\!\pm\!0.5$ \\
 & + MVT (adaptive) &  &  &  &  & $\mathbf{56.9}\!\pm\!0.5$ & $\mathbf{16.6}\!\pm\!2.1$ & $\mathbf{66.1}\!\pm\!0.4$ & $\mathbf{99.5}\!\pm\!0.1$ & $\mathbf{80.4}\!\pm\!1.1$ & $7.9\!\pm\!0.3$ \\
\midrule
Qwen2.5-32B & Text-only & 0.25 & 8.7 & 0.06 & 2.5 & $38.1\!\pm\!0.3$ & $3.9\!\pm\!4.3$ & $61.7\!\pm\!0.9$ & $88.0\!\pm\!1.7$ & $50.7\!\pm\!1.2$ & $\mathbf{15.3}\!\pm\!2.0$ \\
 & + MVT (adaptive) &  &  &  &  & $\mathbf{52.1}\!\pm\!0.8$ & $\mathbf{14.5}\!\pm\!15.9$ & $\mathbf{64.0}\!\pm\!1.2$ & $\mathbf{99.1}\!\pm\!0.1$ & $\mathbf{84.7}\!\pm\!1.2$ & $11.8\!\pm\!0.9$ \\
\midrule
\textit{Average} & Text-only & \multicolumn{4}{c}{---} & $40.4$ & $15.7$ & $61.5$ & $95.5$ & $62.9$ & $\mathbf{14.7}$ \\
 & + MVT (adaptive) & \multicolumn{4}{c}{---} & $\mathbf{57.7}$ & $\mathbf{27.2}$ & $\mathbf{64.0}$ & $\mathbf{99.5}$ & $\mathbf{86.6}$ & $13.4$ \\
\bottomrule
\end{tabular*}

\caption{\textbf{MVT on text-only LLMs.}
Alignment reports held-out NLL (nats) and character error rate
(CER, \%) on 1{,}000 WikiText sentences before MVT training
(lower is better). Task scores compare text-only training
with adaptive MVT. Task scores are percentages;
bold marks the better mean for each model.
The final two rows average task-score means across all five models.}
\label{tab:archagnostic}
\end{table*}

We first evaluate MVT on three tasks using Qwen3-Omni,
then extend it to text-only LLMs. We also test whether
the improvements transfer across input modalities and
whether the observed behavior supports our theoretical
analysis.

\subsection{MVT improves learning, forgetting, and editing}
\label{sec:results-main}
\label{sec:results-qa}
\label{sec:results-unlearn}
\label{sec:results-edit}

Table~\ref{tab:main} compares text-only training and three
text augmentation methods, each alone and with uniform or
adaptive MVT. Compared with text-only training, adaptive
MVT improves QA recall by \gainTWQA{} points, forget quality by
\gainTWFQ{} points, and editing generality by \gainTWGen{} points.
Across the six metrics, uniform and adaptive MVT improve
the average score by \gainTMAvg{} and \gainTWAvg{} points, respectively.

\paragraph{MVT complements text augmentation.}
Adding MVT improves QA recall and editing generality for
paraphrasing, NEFTune, and SymNoise. Paraphrasing is the strongest text
baseline for QA and editing generality, yet adaptive MVT
improves both further, achieving the highest QA recall
(73.9\%). SymNoise with adaptive MVT achieves the highest
forgetting quality (71.3\%). These results show that modality
views provide benefits beyond existing text augmentations.

It is to be noted that we observe absurdly large forget-quality gains with MVT over any baselines. However, this is because the forget quality is a KS-test \(p\)-value, so even a small shift in the truth-ratio distributions can produce a large score increase, as shown in~\cite{eldan2023tofu}. Additionally, locality does not improve with MVT, likely because our editing objective
prioritizes edited facts over preserving unrelated knowledge.
Since MVT is algorithm agnostic, locality-preserving
objectives~\cite{mitchell2022fast} may mitigate this drop.

\pgfplotstableread{
p base baseerr mvt mvterr nb nm
0 45.1 0.9 60.5 0.0 3 1
1 36.7 0.3 59.6 0.4 3 3
2 32.0 0.4 56.6 0.7 3 3
}\pwqa
\pgfplotstableread{
p base baseerr mvt mvterr nb nm
0 7.6 2.6 56.3 17.1 3 3
1 4.1 0.9 61.7 7.0 3 3
2 1.8 0.3 11.3 5.9 3 3
}\pwunlearn
\pgfplotstableread{
p base baseerr mvt mvterr nb nm
0 96.0 0.1 99.6 0.0 3 3
1 36.4 1.7 99.7 0.0 3 3
2 6.8 0.5 99.7 0.1 3 3
}\pwedit
\begin{figure}[t]
\centering
\begin{tikzpicture}
\pgfplotsset{pw/.style={width=2.4cm,height=2.3cm,scale only axis=true,ybar,bar width=4.5pt,
  ymin=0,ymax=105,ytick={0,50,100},xmin=-0.55,xmax=2.55,xtick={0,1,2},xticklabels={Text,Speech,Image},
  axis x line*=bottom,axis y line*=left,tick align=outside,major tick length=2pt,
  tick label style={font=\tiny},title style={font=\scriptsize,yshift=-4pt},ymajorgrids,grid style={black!8},
  enlarge x limits=false}}
\begin{groupplot}[group style={group size=3 by 1,horizontal sep=0.22cm,yticklabels at=edge left},pw,
  legend columns=2,legend style={font=\scriptsize,draw=none,fill=none,column sep=4pt,inner sep=0pt},
  legend cell align=left,legend image code/.code={\draw[#1] (0cm,-0.08cm) rectangle (0.22cm,0.1cm);}]
\nextgroupplot[title={QA recall $\uparrow$},legend to name=pwlegend]
\addplot[draw=none,fill=black!35,error bars/.cd,y dir=both,y explicit,error bar style={line width=0.5pt,black!80},error mark options={rotate=90,mark size=1.3pt,line width=0.5pt}] table[x=p,y=base,y error=baseerr]{\pwqa};
\addlegendentry{Baselines (avg.)}
\addplot[draw=none,fill=viewpurple,error bars/.cd,y dir=both,y explicit,error bar style={line width=0.5pt,black!80},error mark options={rotate=90,mark size=1.3pt,line width=0.5pt}] table[x=p,y=mvt,y error=mvterr]{\pwqa};
\addlegendentry{+ adaptive MVT (avg.)}
\nextgroupplot[title={Forget quality $\uparrow$}]
\addplot[draw=none,fill=black!35,error bars/.cd,y dir=both,y explicit,error bar style={line width=0.5pt,black!80},error mark options={rotate=90,mark size=1.3pt,line width=0.5pt}] table[x=p,y=base,y error=baseerr]{\pwunlearn};
\addplot[draw=none,fill=viewpurple,error bars/.cd,y dir=both,y explicit,error bar style={line width=0.5pt,black!80},error mark options={rotate=90,mark size=1.3pt,line width=0.5pt}] table[x=p,y=mvt,y error=mvterr]{\pwunlearn};
\nextgroupplot[title={Edit reliability $\uparrow$}]
\addplot[draw=none,fill=black!35,error bars/.cd,y dir=both,y explicit,error bar style={line width=0.5pt,black!80},error mark options={rotate=90,mark size=1.3pt,line width=0.5pt}] table[x=p,y=base,y error=baseerr]{\pwedit};
\addplot[draw=none,fill=viewpurple,error bars/.cd,y dir=both,y explicit,error bar style={line width=0.5pt,black!80},error mark options={rotate=90,mark size=1.3pt,line width=0.5pt}] table[x=p,y=mvt,y error=mvterr]{\pwedit};
\end{groupplot}
\node[anchor=south] at ($(group c2r1.north)+(0,0.38)$) {\ref*{pwlegend}};
\end{tikzpicture}
\caption{\textbf{Performance across text, speech, and image queries.}
Grey bars average the four baselines (text-only, paraphrasing, NEFTune, SymNoise); purple bars
average their combinations with adaptive MVT.
Appendix~\ref{sec:appendix-crossmodal} reports all values.}
\label{fig:pathway}
\end{figure}
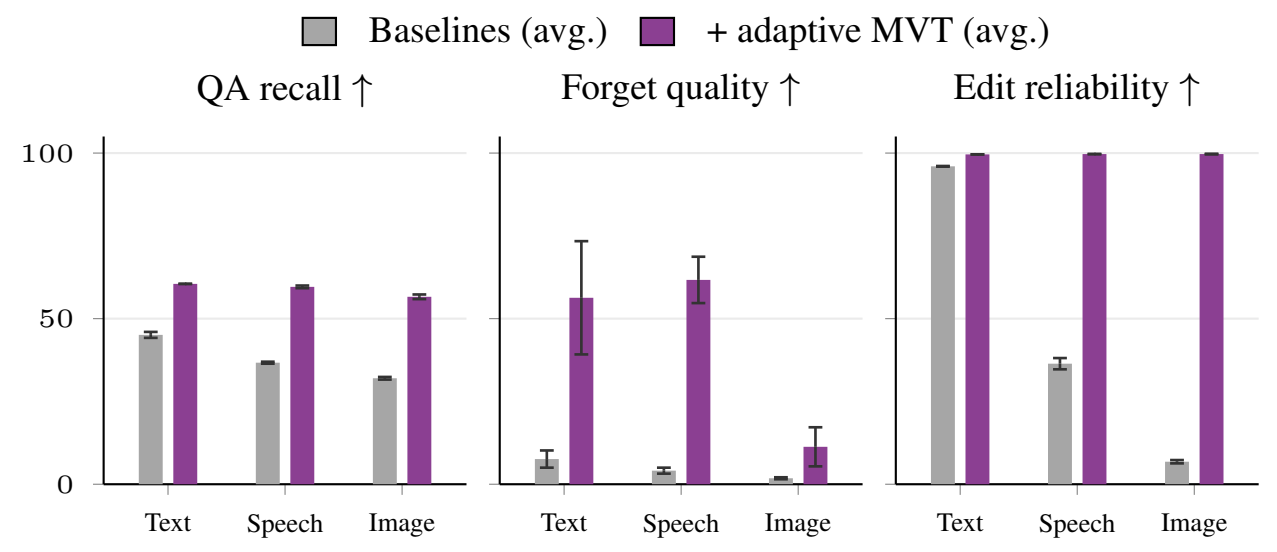
\input{figs/view_agreement_qa}

Table~\ref{tab:qualitative} shows representative examples.
MVT suppresses a targeted author fact revealed by the text
baselines, correctly answers a paraphrased QA question,
and produces the replacement answer for a paraphrased
editing query.

\subsection{MVT extends to text-only LLMs}
\label{sec:arch-agnostic}

We next test MVT on four text-only LLMs using the aligned
modality connectors described in Section~\ref{sec:anybackbone}
(Table~\ref{tab:archagnostic}).

\paragraph{Connector alignment.}
Across the four backbones, held-out text reconstruction
has a mean character error rate of 3.7\% for images and
10.6\% for speech. Thus, the aligned connectors convey the 
source text with some errors, with speech noisier than images.

\paragraph{Downstream performance.}
Adaptive MVT improves QA recall and
editing generality on all four backbones, by \BackboneQAGain{} and
\BackboneEditGain{} points on average. Reliability and model utility
also improve. Mean forget quality increases on Llama, OLMo,
and Qwen and stays the same on Yi, yielding an average gain of
\BackboneForgetGain{} points.
Locality remains a limitation as
discussed in Section~\ref{sec:results-main} Overall, MVT improves
performance even in models without native
multimodal pathways.

\subsection{MVT improves performance across modalities}
\label{sec:crossmodal-editing}

Although MVT targets text-only deployment, evaluating the
trained model through different input pathways reveals
whether the learned changes transfer across modalities.
This is particularly important for unlearning and editing,
where a fact removed or updated through text may remain
accessible through speech or images. We evaluate the
training methods from Table~\ref{tab:main} through all three
pathways and compare the average of the four
baselines (i.e., text-only, paraphrasing, NEFTune, SymNoise) with the average of their adaptive MVT
versions (Figure~\ref{fig:pathway}).

The baselines perform substantially worse on speech and images than on text, particularly for knowledge editing, where reliability drops from 96.0\% on text to 37.5\% on speech and 7.4\% on images. Adaptive MVT largely closes this gap, achieving over 99\% editing reliability and approximately 60\% QA recall across all three pathways. Forget quality also improves substantially on text and speech, although image forgetting remains challenging despite improvements from \cmBaseFQImage{} to \cmMVTFQImage{}.

These improvements require no additional task facts or
encoder updates. By training the shared LLM through
multiple pathways, MVT improves performance not only
on text inputs but also through the auxiliary modalities.

\subsection{Quantifying the theoretical assumptions}
\label{sec:theorytests}

We next investigate whether the observed behavior is
consistent with our theoretical analysis
(Section~\ref{sec:theory}). We examine gradient similarity
across views, disagreement between their predictions,
and similarity of internal activations.

\paragraph{Shared and view-specific gradients.}
Assumption~\ref{ass:S} models each view's gradient as a
shared content direction plus a view-specific surface
direction. To examine this assumption, we compute cosine similarity between each
view's gradient and that of a reworded text question
about the same fact. As a baseline, we compare it with
the gradient of a reworded question about a different
fact. We do this for a mid-training adaptive MVT checkpoint. Figure~\ref{fig:viewagreement}a shows that gradients
associated with the same fact are substantially more
aligned than those associated with different facts
(mean similarity 0.24 versus 0.01). This is
consistent with a shared content component
in Theorem~\ref{thm:diversity}. Additionally, the near-zero random baseline supports
the approximation that gradients of unrelated facts
are weakly aligned.

\paragraph{MVT reduces prediction disagreement.}
Proposition~\ref{prop:agreement} decomposes the weighted
cross-entropy across views into the loss of a combined
prediction and a disagreement term. The combined prediction is obtained by averaging view logits before the softmax, while disagreement is the mean KL divergence from this prediction to each view’s prediction. To examine this proposition, we measure disagreement on 100 training questions of TOFU QA
throughout text-only and adaptive MVT training
(Figure~\ref{fig:viewagreement}b). Under adaptive MVT,
disagreement decreases from 0.245 to 0.001, whereas
it increases under text-only training. This suggests
that jointly training the pathways encourages
consistent predictions across views, while text-only
training allows their predictions to diverge.

\paragraph{Internal representation similarity.}
We further examine whether MVT encourages similar
internal representations across views. At each
evaluated layer, we measure the cosine similarity between
LoRA adapter activations from different views of the
same question, subtracting the similarity between
unrelated questions as a baseline. Figure~\ref{fig:viewagreement}c shows that adaptive
MVT produces higher cosine similarity than text-only
training at all eight evaluated layers. This provides
additional evidence that MVT encourages more similar
internal representations across views. Appendix~\ref{sec:appendix-sas} provides the full
metric definition and view combinations.

Detailed plots across QA, unlearning, and editing, 
tasks and different training runs are provided in
Appendix~\ref{sec:appendix-viewagreement}.

\section{Related Work}
\label{sec:related}

\paragraph{Data augmentation for language model training.}
Text augmentation creates different versions of a training
example while preserving its meaning. Common methods include
word replacement~\cite{wei2019eda},
back-translation~\cite{xie2019uda}, and
paraphrasing~\cite{najafi2024riff,ovadia2024finetuning,
maini2024wrap}.
Contextual augmentation~\cite{kobayashi2018contextual}
replaces words based on their surrounding context.
Other methods generate new examples, as in
Self-Instruct~\cite{wang2023selfinstruct} and
EntiGraph~\cite{yang2025entigraph}.
Instead of changing the words,
MixText~\cite{chen2020mixtext} interpolates hidden
representations, while
NEFTune~\cite{jain2024neftune} and
SymNoise~\cite{yadav2025symnoise} add noise to embeddings.
MVT takes a different approach. It presents the same
example as text, speech, and a printed-text image.
All views use the same supervision to train a shared LLM.
Our adaptive weighting is related to robust optimization,
including Group DRO~\cite{sagawa2020groupdro} and
DoReMi~\cite{xie2023doremi}, giving more
weight to higher-loss modalities.

\paragraph{Multi-view and multimodal learning.}
Multi-view learning~\cite{xu2013multiview} uses different
representations of the same example. Some methods
encourage agreement between views, as in
co-training~\cite{blum1998cotraining}. Others learn
representations that are correlated across views,
as in deep CCA~\cite{andrew2013dcca}.
Multimodal learning can use one modality to improve
learning in another. For example, audio can improve
lip reading~\cite{ngiam2011multimodal}, and RGB images
can help train depth models~\cite{gupta2016crossmodal}.
Visual information has also been used to improve text
models~\cite{tan2020vokenization}.
Cross-modal distillation methods transfer knowledge
between modalities~\cite{xue2023modality,huo2024c2kd}.
Related work also studies how auxiliary or unpaired
modality data can help~\cite{huh2024platonic}. 

There are two closely related methods to MVT: 
Multimodal Pathway~\cite{zhang2024mpathway} connects transformers trained on different modalities and transfers knowledge through cross-modal weight reparameterization. In contrast, MVT uses different input pathways to present the same content and update a shared LLM. Text-Printed Image~\cite{yamabe2026tpi} renders text as images during training to improve visual QA. MVT instead combines text, speech, and printed-image views to improve text-based learning without additional facts or labels.

\paragraph{Unlearning and editing.}
Existing machine unlearning methods use gradient-based
losses~\cite{jang2023knowledge,eldan2023tofu},
preference optimization~\cite{zhang2024npo,fan2024simnpo},
or changes to hidden representations~\cite{li2024wmdp}.
Knowledge editing instead changes specific facts in
a model. Methods include
fine-tuning~\cite{zhu2020modifying},
learned editors~\cite{decao2021editing,mitchell2022fast},
direct weight updates~\cite{meng2022rome,meng2023memit,
fang2025alphaedit},
external memories~\cite{mitchell2022serac,wang2024wise},
and forgetting old facts before learning new
ones~\cite{ni2024flearning}.
Recent studies also test whether forgetting
~\cite{patil2025unlok} and editing
~\cite{zhang2025mcmke,yang2025sake}
work across image and speech inputs.
MVT does not introduce a new unlearning or editing
algorithm. Instead, it applies the same task loss
through different modality views. It may therefore
be easily combined with any unlearning and editing methods.

\section{Conclusion}
\label{sec:conclusion}

We introduced Multimodal View Training (MVT), a simple approach
that uses speech and printed-text images as additional training
views to improve text-based learning. By presenting the same
information through different modality pathways, MVT provides
diverse training signals without requiring additional task
annotations. Our theoretical analysis suggests that training
across modalities can favor updates that capture shared factual
content rather than presentation-specific features. Our
experiments demonstrate improvements across question answering,
machine unlearning, and knowledge editing. We further showed
that MVT complements existing text augmentation methods and
extends to text-only LLMs using lightweight modality connectors
that can be discarded after training. 

Future work will explore
combining MVT with stronger unlearning and editing algorithms,
generating more varied views through different fonts and voices,
and investigating alternative ways of combining views during
training.

\paragraph{Potential negative societal impacts.}
MVT is a general training method whose risks depend on its use. Facts removed or edited through text may still be accessible through
speech or image queries in a multimodal model. MVT reduces this gap but does not fully
resolve it, especially for image forgetting. This is particularly concerning for
personal data, since hiding an answer does not guarantee its removal
from model weights. Because MVT makes
learning more effective, it can also memorize private data more
strongly. Unlearning and editing could be misused to censor information or introduce false or biased facts. Deployments should use authorized data, restrict and log edits,
and audit model behavior across modalities.
\section*{LLM Usage Statement}
We used Claude Code and OpenAI Codex to assist with writing code for experiments, revising the manuscript to fix grammar errors,
reference checks, and figure creation and plotting code. We use ChatGPT to find papers for the related work section of the paper.
The authors retain responsibility for the methods, results, citations,
and final manuscript.
\section*{Acknowledgments}
\addcontentsline{toc}{section}{Acknowledgments}
We acknowledge the NYU HPC Cluster, on which all experiments were run.
Niv Cohen is a Jacques Lewiner Career Advancement Chair Fellow at the Technion.

\clearpage
\phantomsection
\addcontentsline{toc}{section}{References}
{
    \small
    \bibliographystyle{ieeenat_fullname}
    \bibliography{main}
}

\clearpage
\appendix
\section*{Appendix}
\addcontentsline{toc}{section}{Appendix}
\setcounter{table}{0}
\setcounter{figure}{0}
\renewcommand{\thetable}{S\arabic{table}}
\renewcommand{\thefigure}{S\arabic{figure}}
\renewcommand{\theHtable}{appendix.\arabic{table}}
\renewcommand{\theHfigure}{appendix.\arabic{figure}}
\section{Setup and proofs for Section~\ref{sec:theory}}
\label{sec:appendix-theory}

This appendix develops the analysis behind the two results in
Section~\ref{sec:theory}. We study why training on multiple views of the
same fact can help a model answer paraphrased questions. Our approach
is to separate the part of a weight update that reinforces shared factual
content from the part tied to a question's wording or modality. We first
establish how these components affect learning under text-only training,
giving us a basis for understanding what additional views can change.
We then show when training across views places more of the update along
shared content, allowing more of the improvement to carry over when the
question is paraphrased. Finally, we prove
Proposition~\ref{prop:agreement}, explaining how the MVT loss encourages
the views to predict similar answer probabilities.

\paragraph{How weight updates affect the correct answer.}
We measure learning by how much fine-tuning increases the model's log
probability of the correct answer. For a question $x$ with correct answer
$y=(y_1,\dots,y_L)$, let $z_m$ be the input representation for view $m$
from Section~\ref{sec:setup}. We use $\omega$ for the LoRA parameters and
call the log probability of the full correct answer the \emph{answer score}:
\begin{equation}
h(z_m;\omega)=\sum_{t=1}^{L}\log p_\omega(y_t\mid z_m,y_{<t}).
\end{equation}
Under teacher forcing, each prediction is conditioned on the preceding
correct answer tokens $y_{<t}$. The negative of the answer score is the
cross-entropy loss summed over the answer tokens.

To understand why training across views can favor shared factual content,
we study how the views' gradients shape the weight update. We evaluate
those gradients at initialization and hold them fixed, motivated by
evidence that gradient features can remain approximately stable in some
fine-tuning settings~\citep{malladi2023kernel}.

Let $\winit$ be the LoRA parameter vector before fine-tuning. At any
point during or after fine-tuning, the parameters are
$\omega=\winit+\delta$, where $\delta$ is the total weight update since
initialization. For a view with input representation $z$, we call the
gradient of the answer score with respect to the LoRA parameters,
evaluated at initialization, the \emph{gradient feature} $u(z)$:
\begin{equation}
u(z)=\left.\nabla_\omega h(z;\omega)\right|_{\omega=\winit}.
\end{equation}
The first-order Taylor approximation near $\winit$ is
\begin{equation}
h(z;\winit+\delta)-h(z;\winit)\approx u(z)^\top\delta.
\end{equation}
The inner product $u(z)^\top\delta$ predicts how much the weight update
raises or lowers the answer score. The approximation is local because it omits
higher-order terms, which can become significant as fine-tuning moves the
weights away from initialization.

In the analysis below, we use $u(z)^\top\delta$ as the answer score
increase in the fixed-gradient approximation.

Let $\mathcal V$ be the set of views used for training, always including
text. Text-only training uses $\mathcal V=\{T\}$, and full MVT uses all
three views ($\mathcal V=\mathcal M$).
We refer to each training question-answer pair $(x_i,y_i)$ as fact $i$.
To present fact $i$ in view $m$, we encode its question as
$z_{i,m}=\phi_m(r_m(x_i))$ and keep $y_i$ as the target answer.
We write $u_{i,m}=u(z_{i,m})$ for the gradient feature of that answer
in view $m\in\mathcal V$. For the main result, we set a target increase of
$\gamma_i>0$ in the answer score for every training view of pair $i$.
In the linearized model, a weight update meets this target on view $m$
when $u_{i,m}^\top\delta=\gamma_i$. We also consider targets
$\gamma_{i,m}$ that can differ across views $m$.

\paragraph{Shared and view-specific gradient components (Assumption~\ref{ass:S}).}
We write the gradient feature for fact $i$ in view $m$ as
$u_{i,m}=a_i+e_{i,m}$. The \emph{content direction} $a_i$ ($\|a_i\|=1$)
is shared by all views. The \emph{surface direction} $e_{i,m}$ depends
on the view and the wording, with surface energy
$\sigma_{i,m}^2=\|e_{i,m}\|^2$. Every content direction is orthogonal to
every surface direction ($a_i^\top e_{j,n}=0$ for all $i,j,n$).
Directions associated with different facts are also orthogonal
($a_i^\top a_j=0$ and $e_{i,m}^\top e_{j,n}=0$ for $i\ne j$ and all $m,n$).
These assumptions imply that the gradient features of different facts
are orthogonal ($u_{i,m}^{\top}u_{j,n}=0$ for $i\ne j$).
Surface directions ($e_{i,m}$) from different views of the same fact
can have nonzero dot products. Their dot products form the surface Gram matrix
$G_i$, with $(G_i)_{mn}=e_{i,m}^\top e_{i,n}$ for
$m,n\in\mathcal V$. We assume $G_i$ is positive definite.

A paraphrased text question keeps the content direction but changes the wording.
We model its gradient feature as $u'_i=a_i+\rho_i e_{i,T}+\xi_i$.
Here $\rho_i\in[0,1]$ measures how much of the original text surface direction
the paraphrased question keeps, and $\xi_i$ represents cues introduced by the
new wording and is orthogonal to all training gradient features.
\paragraph{Transfer ratio.}
The \emph{transfer ratio} $\nu_i=u_i'^\top\delta/\gamma_i$ measures how much
of the target score increase $\gamma_i$ is preserved after paraphrasing.
For both text-only training and MVT, we measure transfer from the original
text question to its paraphrased text version.

\subsection{Gradient descent and the minimum-norm weight update}

Eq.~\eqref{eq:lin} approximates the answer score increase for fact $i$
in view $m$ by $u_{i,m}^{\top}\delta$. For Lemma~\ref{lem:span}, we
consider training objectives in which the loss for predicting answer
$y_i$ from this view depends on $\delta$ through that score increase.
Cross-entropy provides one example. Since it is the negative answer
score, its value under Eq.~\eqref{eq:lin} is
\begin{equation}
\ell_{i,m}(u_{i,m}^{\top}\delta)
=-h(z_{i,m};\winit)-u_{i,m}^{\top}\delta.
\end{equation}
Here, $\ell_{i,m}(s)=-h(z_{i,m};\winit)-s$ takes a predicted score increase
$s$ and returns the corresponding loss. The initial score is fixed for
this training example, so it enters the function as a constant.
More generally, for each fact $i$ and training view $m$, we assume that
the loss depends on the weight update through the predicted answer score
increase:
\begin{equation}\label{eq:supp-loss-form}
\text{loss for fact $i$ in view $m$}
=\ell_{i,m}(u_{i,m}^{\top}\delta).
\end{equation}
Here, $\ell_{i,m}$ is differentiable and the initial score is held fixed.
The cross-entropy example above has this form.

\begin{lemma}[Gradient descent produces combinations of the training gradient features]\label{lem:span}
Starting from the initial LoRA parameters $\winit$, gradient descent
or SGD produces an accumulated weight update
$\delta_t=\omega_t-\winit$ that is a linear combination of the training
gradient features $u_{i,m}$ at every step $t$. This holds for any minibatch
sequence, learning rates, and positive loss weights, assuming each loss
depends on $\delta$ through $u_{i,m}^{\top}\delta$, as specified in
Eq.~\eqref{eq:supp-loss-form}.

Under Assumption~\ref{ass:S}, the component of $\delta_t$ along the content
and surface directions of fact $i$ has the form $b_{i,t}(a_i+e_{i,T})$
during text-only training, where $b_{i,t}$ is a scalar.

Furthermore, let $\delta^\star$ be the update at any training step that
achieves the specified answer score increases on all training facts and
views:
\begin{equation}\label{eq:supp-score-constraints}
u_{i,m}^{\top}\delta^\star=\gamma_{i,m}
\qquad\text{for every training $i,m$}.
\end{equation}
Then $\delta^\star$ has the smallest Euclidean norm among all weight
updates satisfying Eq.~\eqref{eq:supp-score-constraints}.
\end{lemma}

\begin{proof}
Let $c_{i,m}>0$ be the weight assigned to the loss for fact $i$ in
view $m$. Summing over the training facts and their views gives the
total training objective
\begin{equation}
\mathcal L(\delta)=\sum_i\sum_{m\in\mathcal V} c_{i,m}\,\ell_{i,m}(u_{i,m}^\top\delta).
\end{equation}
To find the gradient of the training objective, we differentiate each
loss with respect to $\delta$. For fact $i$ in view $m$, the loss is
$\ell_{i,m}(u_{i,m}^{\top}\delta)$. Applying the chain rule of
differentiation gives
\begin{equation}\label{eq:supp-loss-gradient}
\nabla_\delta\ell_{i,m}(u_{i,m}^{\top}\delta)
=\ell_{i,m}'(u_{i,m}^{\top}\delta)\,u_{i,m}.
\end{equation}
Here, $\ell_{i,m}'(u_{i,m}^{\top}\delta)$ is a scalar, and $u_{i,m}$ is
fixed in our approximation. Including the loss weight $c_{i,m}$, the
contribution of this fact and view to the total gradient is therefore
a scalar multiple of $u_{i,m}$. Summing these contributions gives
\begin{equation}\label{eq:supp-objective-gradient}
\nabla_\delta\mathcal L(\delta)
=\sum_i\sum_{m\in\mathcal V}
\underbrace{c_{i,m}\ell_{i,m}'(u_{i,m}^{\top}\delta)}_{\text{scalar coefficient}}
u_{i,m}.
\end{equation}
This is a linear combination of the training gradient features:
each $u_{i,m}$ is multiplied by a scalar, and the resulting vectors
are added. A minibatch contains a subset of the training facts and
views. Its gradient sums the corresponding terms, so it is also a
linear combination of the training gradient features.

Let $\mathcal S=\operatorname{span}\{u_{i,m}\}$ be the set of all linear
combinations of the training gradient features. Before training, the LoRA
parameters satisfy $\omega_0=\winit$, so the accumulated weight update is
$\delta_0=0$. Each gradient step adds a linear combination of the $u_{i,m}$ to
the accumulated update. Starting from zero, every $\delta_t$ therefore
belongs to $\mathcal S$.

For text-only training, the gradient feature for fact $i$ is
$u_{i,T}=a_i+e_{i,T}$. Under Assumption~\ref{ass:S}, features associated with
other facts are perpendicular to both $a_i$ and $e_{i,T}$. The components
of the weight update associated with other facts therefore contribute
zero to the answer score increase for fact $i$. The component associated
with fact $i$ is a scalar multiple $b_{i,t}(a_i+e_{i,T})$.

To establish the minimum-norm claim, consider any update $\delta$
satisfying Eq.~\eqref{eq:supp-score-constraints}. Decompose it into its
projection $\delta_{\mathcal S}$ onto $\mathcal S$ and a perpendicular
component $\delta_\perp$. Since $\delta_\perp$ is perpendicular to every
$u_{i,m}$,
\begin{gather}
u_{i,m}^\top\delta=u_{i,m}^\top\delta_{\mathcal S},\\
\|\delta\|_2^2
=\|\delta_{\mathcal S}\|_2^2+\|\delta_\perp\|_2^2.
\end{gather}
Removing $\delta_\perp$ preserves every required score increase and
reduces the update's norm whenever $\delta_\perp\neq0$. Therefore, the
weight update with the smallest Euclidean norm satisfying
Eq.~\eqref{eq:supp-score-constraints} must lie entirely in $\mathcal S$.
To show that there is at most one update in $\mathcal S$ satisfying these
constraints, suppose two such updates exist and let $d$ be their
difference. Subtracting their score-increase equations gives
$u_{i,m}^{\top}d=0$ for every training gradient feature. Since $d$ belongs
to $\mathcal S$, it is a linear combination of those features and must
therefore be perpendicular to itself. Thus $d^\top d=\|d\|_2^2=0$,
implying $d=0$, so the two updates are identical.

By the span argument above, $\delta^\star$ belongs to $\mathcal S$.
The uniqueness argument therefore identifies it as the minimum-norm
update satisfying Eq.~\eqref{eq:supp-score-constraints}. This conclusion
applies at any training step achieving those score increases.
\end{proof}

\subsection{Text-only transfer}
\label{sec:appendix-textonly}

Under text-only training, Lemma~\ref{lem:span} shows that the component
of the weight update associated with fact $i$ is $b_{i,t}(a_i+e_{i,T})$.
At every training step, this component remains a scalar multiple of the
same vector $a_i+e_{i,T}$. Only the scalar $b_{i,t}$ can change.
Corollary~\ref{cor:textonly} shows how this fixed relationship determines
transfer from the original text question to a paraphrased question during
text-only training.

\begin{corollary}[Text-only transfer is fixed]\label{cor:textonly}
Let $\sigma_{i,T}^2=\|e_{i,T}\|_2^2$. At training step $t$, define the
transfer ratio as the answer score increase on the paraphrased question
divided by the increase on the original text question:
\begin{equation}
\nu_i(t)=\frac{u_i'^{\top}\delta_t}{u_{i,T}^{\top}\delta_t},
\qquad u_{i,T}^{\top}\delta_t\neq0.
\end{equation}
Under the assumptions of Lemma~\ref{lem:span} and Assumption~\ref{ass:S},
\begin{equation}\label{eq:supp-textonly-shortfall}
1-\nu_i(t)=\frac{(1-\rho_i)\sigma_{i,T}^2}{1+\sigma_{i,T}^2}.
\end{equation}
This ratio is independent of $t$: text-only training preserves the
relative gap between improvement on the training question and its
paraphrase. If training produces the specified increase $\gamma_i$ on the original
text question, then $u_{i,T}^{\top}\delta_t=\gamma_i$, and this ratio equals
$\nu_i$ defined in Section~\ref{sec:theory-transfer}.
\end{corollary}

\begin{proof}
By Lemma~\ref{lem:span}, the component of $\delta_t$ associated with fact
$i$ is $b_{i,t}(a_i+e_{i,T})$. Under Assumption~\ref{ass:S}, components
associated with other facts contribute zero to the answer score
increases on both questions.

For the original text question, $u_{i,T}=a_i+e_{i,T}$. Substituting this
gradient feature and the fact-$i$ component of the update into
Eq.~\eqref{eq:lin} gives
\begin{equation}
\begin{aligned}
u_{i,T}^{\top}\delta_t
&=(a_i+e_{i,T})^\top b_{i,t}(a_i+e_{i,T})\\
&=b_{i,t}\|a_i+e_{i,T}\|_2^2\\
&=b_{i,t}(1+\sigma_{i,T}^2).
\end{aligned}
\end{equation}
The last equality follows from Assumption~\ref{ass:S}, which gives
$\|a_i\|_2=1$, $a_i^\top e_{i,T}=0$, and
$\sigma_{i,T}^2=\|e_{i,T}\|_2^2$.

For the paraphrased question, we use its gradient feature
$u_i'=a_i+\rho_i e_{i,T}+\xi_i$ in the same calculation:
\begin{equation}
\begin{aligned}
u_i'^{\top}\delta_t
&=(a_i+\rho_i e_{i,T}+\xi_i)^\top
 b_{i,t}(a_i+e_{i,T})\\
&=b_{i,t}(1+\rho_i\sigma_{i,T}^2).
\end{aligned}
\end{equation}
Here, the orthogonality conditions in Assumption~\ref{ass:S} remove
the cross terms, including the contribution from $\xi_i$. Dividing the two score increases cancels $b_{i,t}$, giving
\begin{equation}
\nu_i(t)=\frac{1+\rho_i\sigma_{i,T}^2}{1+\sigma_{i,T}^2}.
\end{equation}
Rearranging gives Eq.~\eqref{eq:supp-textonly-shortfall}, which holds at
every training step where $\nu_i(t)$ is defined. In particular, when the
original text question's score increase equals $\gamma_i$, the ratio
becomes
\begin{equation}
\nu_i(t)=\frac{u_i'^\top\delta_t}{\gamma_i}=\nu_i,
\end{equation}
using the definition in Section~\ref{sec:theory-transfer}. The corollary
therefore gives the transfer ratio for text-only training as
\begin{equation}\label{eq:supp-textonly-transfer}
\nu_i=1-\frac{(1-\rho_i)\sigma_{i,T}^2}{1+\sigma_{i,T}^2}.
\end{equation}
\end{proof}

\subsection{Proof of Theorem~\ref{thm:diversity}}

Lemma~\ref{lem:span} establishes that a gradient-descent or SGD update
achieving the specified score increases has minimum Euclidean norm.
We therefore calculate that minimum-norm update. The lemma also
shows that the update ($\delta$) is a linear combination of the training gradient
features. Under Assumption~\ref{ass:S}, we can separate it into orthogonal
components associated with individual facts:
\begin{equation}\label{eq:supp-fact-decomposition}
\delta=\sum_i\delta_i,
\qquad
\delta_i^\top\delta_j=0\quad(i\neq j).
\end{equation}
Here, $\delta_i$ is the component associated with fact $i$. These
components together form the single weight update applied to the model.
Orthogonality gives
\begin{equation}\label{eq:supp-fact-norms}
\|\delta\|_2^2=\sum_i\|\delta_i\|_2^2,
\qquad
u_{i,m}^\top\delta=u_{i,m}^\top\delta_i.
\end{equation}
Each fact's score increase depends only on its own component
$(\delta_i)$. Since $\|\delta\|_2^2=\sum_i\|\delta_i\|_2^2$, the minimum-norm
update $\delta^\star$ must contain a minimum-norm component
$\delta_i^\star$ for each fact $i$. Otherwise, replacing one component
with a smaller one satisfying the same constraints would reduce the
full update's norm while preserving every fact's score increases.
We can therefore find $\delta^\star$ by minimizing each component
separately. For fact $i$, we require
$u_{i,m}^\top\delta_i=\gamma_i$ on every training view $m$.

Let $M_i$ collect the surface directions for fact $i$, with one column
per training view. Text-only training gives $M_i=[e_{i,T}]$, and training
with all three views gives $M_i=[e_{i,T},e_{i,A},e_{i,I}]$. The surface Gram
matrix $G_i$ defined in Assumption~\ref{ass:S} is therefore
$G_i=M_i^\top M_i$. We also use the quantity
$\Pi_i=\one^\top G_i^{-1}\one$ defined in Section~\ref{sec:theory-transfer}.

By Lemma~\ref{lem:span}, the update for fact $i$ is a combination of the
training gradient features $u_{i,m}=a_i+e_{i,m}$. We can therefore write
\begin{equation}\label{eq:supp-content-surface-update}
\delta_i=\alpha a_i+M_i\beta_i.
\end{equation}
Here, the scalar $\alpha$ is the coefficient along the shared content
direction, and the entries of the vector $\beta_i$ are the coefficients
along the surface directions. Because $a_i$ is perpendicular to every
column of $M_i$ and has unit norm, $a_i^\top\delta_i=\alpha$. At the
minimizing update, $\alpha^\star=a_i^\top\delta_i^\star=a_i^\top\delta^\star$,
matching Section~\ref{sec:theory-transfer}.

For training view $m$, the answer score increase is
\begin{equation}\label{eq:supp-view-score-increase}
\begin{aligned}
u_{i,m}^\top\delta_i
&=(a_i+e_{i,m})^\top(\alpha a_i+M_i\beta_i)\\
&=\alpha+e_{i,m}^\top M_i\beta_i\\
&=\alpha+(G_i\beta_i)_m.
\end{aligned}
\end{equation}
The last equality follows because the $m$-th row of $G_i$ contains the
dot products of $e_{i,m}$ with the surface-direction columns of $M_i$.
We require $u_{i,m}^\top\delta_i=\gamma_i$, so
Eq.~\eqref{eq:supp-view-score-increase} must satisfy
$\alpha+(G_i\beta_i)_m=\gamma_i$ for every training view $m$.
Collecting these equations gives
\begin{equation}\label{eq:supp-view-score-constraints}
\alpha\one+G_i\beta_i=\gamma_i\one.
\end{equation}
Subtracting $\alpha\one$ and multiplying by $G_i^{-1}$ gives
\begin{equation}\label{eq:supp-surface-coefficients}
\beta_i=(\gamma_i-\alpha)G_i^{-1}\one.
\end{equation}

We next compute the squared norm of the update in
Eq.~\eqref{eq:supp-content-surface-update} by taking its dot product
with itself:
\begin{equation}\label{eq:supp-update-norm-expansion}
\begin{aligned}
\|\delta_i\|_2^2
&=(\alpha a_i+M_i\beta_i)^\top(\alpha a_i+M_i\beta_i)\\
&=\alpha^2a_i^\top a_i+2\alpha a_i^\top M_i\beta_i\\
&\quad+\beta_i^\top M_i^\top M_i\beta_i\\
&=\alpha^2+\beta_i^\top G_i\beta_i.
\end{aligned}
\end{equation}
The last equality uses $\|a_i\|_2=1$, $a_i^\top M_i=0$, and
$M_i^\top M_i=G_i$. Substituting the expression for $\beta_i$ in
Eq.~\eqref{eq:supp-surface-coefficients}, and using the symmetry of $G_i$,
gives
\begin{equation}\label{eq:supp-update-norm-alpha}
\begin{aligned}
\|\delta_i\|_2^2
&=\alpha^2+(\gamma_i-\alpha)^2\one^\top G_i^{-1}\one\\
&=\alpha^2+\Pi_i(\gamma_i-\alpha)^2.
\end{aligned}
\end{equation}
Substituting the score constraints has expressed the squared update
norm entirely in terms of $\alpha$. Each choice of $\alpha$ determines
the surface coefficients $\beta_i$ through
Eq.~\eqref{eq:supp-surface-coefficients}. We therefore minimize this
expression over $\alpha$ to find the minimum-norm update.
Differentiating gives
\begin{equation}\label{eq:supp-norm-derivative}
\frac{d}{d\alpha}\|\delta_i\|_2^2
=2\alpha-2\Pi_i(\gamma_i-\alpha).
\end{equation}
Setting this derivative to zero and solving for $\alpha$ yields
\begin{equation}\label{eq:supp-optimal-content}
\alpha^\star=\frac{\gamma_i\Pi_i}{1+\Pi_i}.
\end{equation}
The second derivative is $2+2\Pi_i>0$, so this is the unique minimum.

The score constraints in Eq.~\eqref{eq:supp-view-score-constraints}
state that $\alpha\one+G_i\beta_i=\gamma_i\one$. At the minimizing
update, substituting $\alpha^\star$ gives
\begin{equation}\label{eq:supp-optimal-surface}
\begin{aligned}
G_i\beta_i^\star
&=(\gamma_i-\alpha^\star)\one\\
&=\left(\gamma_i-\frac{\gamma_i\Pi_i}{1+\Pi_i}\right)\one\\
&=\frac{\gamma_i}{1+\Pi_i}\one.
\end{aligned}
\end{equation}
Eq.~\eqref{eq:supp-view-score-increase} expresses the answer score
increase on view $m$ as $\alpha^\star+(G_i\beta_i^\star)_m$. Thus,
the shared content component of the weight update raises the answer
score by $\alpha^\star$, and the surface component raises it by
$\gamma_i/(1+\Pi_i)$. These two contributions sum to the required
increase $\gamma_i$ on every training view.

For the paraphrased text question, Assumption~\ref{ass:S} gives the
gradient feature $u_i'=a_i+\rho_i e_{i,T}+\xi_i$. Substituting this feature
and the update decomposition from
Eq.~\eqref{eq:supp-content-surface-update} gives
\begin{equation}\label{eq:supp-paraphrase-score-components}
\begin{aligned}
u_i'^\top\delta_i^\star
&=(a_i+\rho_i e_{i,T}+\xi_i)^\top
(\alpha^\star a_i+M_i\beta_i^\star)\\
&=\alpha^\star+\rho_i e_{i,T}^\top M_i\beta_i^\star
+\xi_i^\top\delta_i^\star\\
&=\alpha^\star+\rho_i(G_i\beta_i^\star)_T.
\end{aligned}
\end{equation}
Here, Assumption~\ref{ass:S} removes the cross terms between content and
surface directions. It also makes $\xi_i$ orthogonal to every training
gradient feature. Since $\delta_i^\star$ is a combination of those
features, $\xi_i^\top\delta_i^\star=0$.

Substituting the expressions for $\alpha^\star$ and
$(G_i\beta_i^\star)_T$, the answer score increase on the paraphrased
question is
\begin{equation}\label{eq:supp-paraphrase-score-gain}
\begin{aligned}
u_i'^\top\delta_i^\star
&=\frac{\gamma_i\Pi_i}{1+\Pi_i}
+\rho_i\frac{\gamma_i}{1+\Pi_i}\\
&=\frac{\gamma_i(\Pi_i+\rho_i)}{1+\Pi_i}.
\end{aligned}
\end{equation}
Dividing by the training increase $\gamma_i$
gives
\begin{equation}
\nu_i=\frac{\Pi_i+\rho_i}{1+\Pi_i}
=1-\frac{1-\rho_i}{1+\Pi_i},
\end{equation}
which proves the transfer formula. For text-only training,
$G_i=[\sigma_{i,T}^2]$, so $\Pi_i=1/\sigma_{i,T}^2$, recovering
Corollary~\ref{cor:textonly}.

We now add one training view of fact $i$ to the existing views. Let
$e_{i,\mathrm{new}}$ be its surface direction and let
$\sigma_{i,\mathrm{new}}^2=\|e_{i,\mathrm{new}}\|_2^2$. The vector
$v_i=M_i^\top e_{i,\mathrm{new}}$ contains the dot products between the
existing surface directions and the new one. Appending the new direction
to $M_i$ gives the enlarged Gram matrix
\begin{equation}\label{eq:supp-enlarged-gram}
G_i'=
\begin{pmatrix}
G_i&v_i\\
v_i^\top&\sigma_{i,\mathrm{new}}^2
\end{pmatrix}.
\end{equation}
The top-left block contains the existing dot products. The last column
and row contain dot products involving the added view, and the
bottom-right entry is its squared surface-direction length.

By definition, the new quantity is
$\Pi_i'=\one^\top(G_i')^{-1}\one$. To calculate it, write
\begin{equation}\label{eq:supp-inverse-gram-vector}
\begin{pmatrix}r_i\\ \tau_i\end{pmatrix}
=(G_i')^{-1}\one.
\end{equation}
This column vector has one entry per training view. The vector $r_i$
contains its entries for the existing views, and the scalar $\tau_i$
is its final entry, corresponding to the added view.

Multiplying both sides by $G_i'$ gives
\begin{equation}\label{eq:supp-added-view-system}
\begin{pmatrix}
G_i&v_i\\
v_i^\top&\sigma_{i,\mathrm{new}}^2
\end{pmatrix}
\begin{pmatrix}r_i\\ \tau_i\end{pmatrix}
=
\begin{pmatrix}\one\\1\end{pmatrix}.
\end{equation}
Reading the top block and bottom row yields
\begin{equation}\label{eq:supp-added-view-top}
G_i r_i+v_i\tau_i=\one,
\end{equation}
\begin{equation}\label{eq:supp-added-view-bottom}
v_i^\top r_i+\sigma_{i,\mathrm{new}}^2\tau_i=1.
\end{equation}
The first equation gives $r_i=G_i^{-1}(\one-v_i\tau_i)$.
Substituting this expression into the second gives
\begin{equation}\label{eq:supp-added-view-elimination}
v_i^\top G_i^{-1}\one
+\bigl(\sigma_{i,\mathrm{new}}^2-v_i^\top G_i^{-1}v_i\bigr)\tau_i=1.
\end{equation}
Solving for $\tau_i$ therefore gives
\begin{equation}\label{eq:supp-added-view-scalar}
\tau_i=\frac{1-v_i^\top G_i^{-1}\one}
{\sigma_{i,\mathrm{new}}^2-v_i^\top G_i^{-1}v_i}.
\end{equation}
The denominator is positive because the enlarged Gram matrix is assumed
positive definite.

Since $\Pi_i'$ is $\one^\top$ multiplied by the column vector
$(G_i')^{-1}\one$, it equals the sum of that vector's entries:
\begin{equation}\label{eq:supp-added-view-entry-sum}
\Pi_i'=\one^\top r_i+\tau_i.
\end{equation}
Substituting $r_i=G_i^{-1}(\one-v_i\tau_i)$ and using
$\Pi_i=\one^\top G_i^{-1}\one$ gives
\begin{equation}\label{eq:supp-added-view-monotonicity}
\begin{aligned}
\Pi_i'
&=\one^\top G_i^{-1}(\one-v_i\tau_i)+\tau_i\\
&=\Pi_i+\bigl(1-v_i^\top G_i^{-1}\one\bigr)\tau_i\\
&=\Pi_i+\frac{\bigl(1-v_i^\top G_i^{-1}\one\bigr)^2}
{\sigma_{i,\mathrm{new}}^2-v_i^\top G_i^{-1}v_i}\\
&\geq\Pi_i.
\end{aligned}
\end{equation}
The second equality uses the symmetry of $G_i^{-1}$. The final inequality
follows because the numerator is a square and the denominator is
positive. This proves that adding the training view cannot decrease
$\Pi_i$. \qed

This analysis applies at any training step achieving the specified
score increases under our fixed-gradient approximation.

\subsection{Unequal score increases across views}

Theorem~\ref{thm:diversity} uses the same answer score increase for every
training view. Here, we allow a different increase for each view. For
example, if we want every view to reach a common final answer score, a
view with a lower initial score needs a larger increase.

For fact $i$, let $\gamma_{i,m}>0$ be the specified score increase for
view $m$, and collect
these increases in the vector $\boldsymbol{\gamma}_i$. We continue to
write the update as $\delta_i=\alpha a_i+M_i\beta_i$, where $\alpha$ is its
shared content component and $\beta_i$ contains its surface coefficients.

The calculation in Appendix~A.3 gives the score increase on view $m$ as
\begin{equation}
u_{i,m}^\top\delta_i=\alpha+(G_i\beta_i)_m.
\end{equation}
Requiring this increase to equal $\gamma_{i,m}$ gives one equation for each
training view. Collecting these equations yields
\begin{equation}
\alpha\one+G_i\beta_i=\boldsymbol{\gamma}_i.
\end{equation}
Thus,
\begin{equation}
\beta_i=G_i^{-1}(\boldsymbol{\gamma}_i-\alpha\one).
\end{equation}

Substituting this expression into the squared update norm gives
\begin{equation}
\|\delta_i\|_2^2
=\alpha^2+(\boldsymbol{\gamma}_i-\alpha\one)^\top
G_i^{-1}(\boldsymbol{\gamma}_i-\alpha\one).
\end{equation}
Differentiating with respect to $\alpha$, the minimum satisfies
\begin{equation}
2\alpha-2\one^\top G_i^{-1}(\boldsymbol{\gamma}_i-\alpha\one)=0.
\end{equation}
Using $\Pi_i=\one^\top G_i^{-1}\one$, we obtain
\begin{equation}\label{eq:supp-unequal-content}
\alpha^\star=\frac{\one^\top G_i^{-1}\boldsymbol{\gamma}_i}{1+\Pi_i}.
\end{equation}
The second derivative is $2(1+\Pi_i)>0$, so this is the unique minimum.

We now calculate transfer to the paraphrased text question. Its gradient
feature is $u_i'=a_i+\rho_i e_{i,T}+\xi_i$, and its score increase is
\begin{equation}\label{eq:supp-unequal-paraphrase-components}
u_i'^\top\delta_i^\star=\alpha^\star+\rho_i(G_i\beta_i^\star)_T.
\end{equation}
The shared content contributes $\alpha^\star$, the original text's
surface component contributes $\rho_i(G_i\beta_i^\star)_T$, and $\xi_i$
contributes zero because it is orthogonal to the training gradient
features.

The constraint for the text view is
\begin{equation}\label{eq:supp-unequal-text-constraint}
\alpha^\star+(G_i\beta_i^\star)_T=\gamma_{i,T}.
\end{equation}
Eq.~\eqref{eq:supp-unequal-text-constraint} gives
$(G_i\beta_i^\star)_T=\gamma_{i,T}-\alpha^\star$. Substituting this into the
paraphrased question's score increase in
Eq.~\eqref{eq:supp-unequal-paraphrase-components} gives
\begin{equation}
u_i'^\top\delta_i^\star=\alpha^\star+\rho_i(\gamma_{i,T}-\alpha^\star).
\end{equation}
Define the transfer ratio $\nu_i$ by dividing this score increase by the
training increase on the text view:
\begin{equation}\label{eq:supp-unequal-transfer}
\nu_i=\frac{u_i'^\top\delta_i^\star}{\gamma_{i,T}}
=\rho_i+(1-\rho_i)\frac{\alpha^\star}{\gamma_{i,T}}.
\end{equation}
Consequently,
\begin{equation}\label{eq:supp-unequal-shortfall}
1-\nu_i=(1-\rho_i)\left(1-\frac{\alpha^\star}{\gamma_{i,T}}\right).
\end{equation}
Eq.~\eqref{eq:supp-unequal-shortfall} extends the transfer formula to
unequal training score increases. For fixed $\gamma_{i,T}$ and $\rho_i$,
increasing the shared content component $\alpha^\star$ can still improve
transfer. Setting every $\gamma_{i,m}=\gamma_i$ recovers
Theorem~\ref{thm:diversity}.

We now compare two training objectives using the same text, audio, and
image views. The first requires the text score increase $\gamma_{i,T}$
on every view. The second allows audio and image to require larger
increases. We examine whether these larger increases strengthen the
shared content component and improve transfer to the paraphrased text
question.

Assume the surface directions are mutually orthogonal, meaning
$e_{i,m}^\top e_{i,n}=0$ for distinct views $m\neq n$. Then $G_i$ is diagonal
with entries $\sigma_{i,m}^2$. Substituting its diagonal inverse into
Eq.~\eqref{eq:supp-unequal-content} gives
\begin{equation}\label{eq:supp-unequal-orthogonal-content}
\alpha^\star=\frac{\sum_m\gamma_{i,m}/\sigma_{i,m}^2}{1+\Pi_i},
\qquad
\Pi_i=\sum_m\frac{1}{\sigma_{i,m}^2}.
\end{equation}
Each view's required increase contributes positively to the numerator.
Increasing the audio or image requirement therefore increases
$\alpha^\star$, while the denominator stays fixed because the training
views and their gradient features are unchanged.

A larger required increase on an auxiliary view makes the shared content
direction more useful: an update along that direction helps meet the
requirements of all views at once. The minimum-norm solution consequently
assigns more of the update to shared content. Because the paraphrased
question shares this content, its answer score also benefits.

Thus, when $\gamma_{i,A},\gamma_{i,I}\geq\gamma_{i,T}$,
\begin{equation}\label{eq:supp-unequal-content-bound}
\alpha^\star\geq\frac{\gamma_{i,T}\Pi_i}{1+\Pi_i}.
\end{equation}
The right side is the shared content component obtained by requiring
$\gamma_{i,T}$ on all three views. The unequal-increase objective therefore
yields at least as much shared content as this equal-increase objective.
Eq.~\eqref{eq:supp-unequal-transfer} then gives a transfer ratio at least
as high as in that equal-increase comparison.

\subsection{Proof of Proposition~\ref{prop:agreement}}

The preceding analysis examines how weight updates affect transfer to
paraphrased questions. We now examine the MVT loss directly.
Proposition~\ref{prop:agreement} shows that the MVT loss splits into a
loss on the combined prediction and a term measuring disagreement across
views.

Consider one token prediction in the answer, with correct token $y$.
For view $m$, the vector $s_m$ contains one logit for each vocabulary
token. We write $s_{m,k}$ for the logit of token $k$, and $s_{m,y}$ for
the logit of the correct token. The predicted probability of token $k$
is
\begin{equation}\label{eq:supp-softmax-probabilities}
p_{m,k}=\frac{\exp(s_{m,k})}{\sum_r\exp(s_{m,r})},
\end{equation}
where the denominator sums over all vocabulary tokens $r$.

Let $w_m\geq0$ be the weight assigned to view $m$, with $\sum_mw_m=1$.
The weighted average of the logit vectors is $\bar s=\sum_mw_ms_m$,
and its softmax gives the combined prediction $\bar p$. In particular,
$\bar s_y=\sum_mw_ms_{m,y}$ is the averaged logit of the correct token.

Eq.~\eqref{eq:agreement} is the likelihood form of the ensemble ambiguity
decomposition~\citep{krogh1995neural,heskes1998bias,wood2023unified}.
We restate it for training across views.

\begin{proof}
For a logit vector $s$, define its log-sum-exp by
\begin{equation}
\mathrm{LSE}(s)=\log\sum_k\exp(s_k).
\end{equation}
Cross-entropy is the negative log probability of the correct token.
Substituting the softmax formula gives
\begin{equation}
\begin{aligned}
\ell_{\mathrm{CE}}(s,y)
&=-\log\frac{\exp(s_y)}{\sum_k\exp(s_k)}\\
&=\log\sum_k\exp(s_k)-s_y\\
&=\mathrm{LSE}(s)-s_y.
\end{aligned}
\end{equation}

We first calculate the difference between the MVT loss and the
cross-entropy of the combined prediction:
\begin{equation}\label{eq:supp-agreement-loss-difference}
\begin{aligned}
&\sum_mw_m\ell_{\mathrm{CE}}(s_m,y)
-\ell_{\mathrm{CE}}(\bar s,y)\\
&=\sum_mw_m\bigl[\mathrm{LSE}(s_m)-s_{m,y}\bigr]\\
&\quad-\bigl[\mathrm{LSE}(\bar s)-\bar s_y\bigr]\\
&=\sum_mw_m\mathrm{LSE}(s_m)-\mathrm{LSE}(\bar s).
\end{aligned}
\end{equation}
The terms involving the correct token sum to zero because
$-\sum_mw_ms_{m,y}+\bar s_y=0$, by the definition of $\bar s_y$.
It remains to show that this difference equals the disagreement term
in Proposition~\ref{prop:agreement}.

For each view $m$, the KL divergence from the combined prediction to
that view's prediction is
\begin{equation}
D_{\mathrm{KL}}(\bar p\|p_m)
=\sum_k\bar p_k\bigl(\log\bar p_k-\log p_{m,k}\bigr).
\end{equation}
Taking logarithms of the softmax formula in
Eq.~\eqref{eq:supp-softmax-probabilities}, applied to $s_m$ and $\bar s$,
gives
\begin{align}
\log\bar p_k&=\bar s_k-\mathrm{LSE}(\bar s),\\
\log p_{m,k}&=s_{m,k}-\mathrm{LSE}(s_m).
\end{align}
Substituting these expressions into the KL divergence yields
\begin{equation}
\begin{aligned}
D_{\mathrm{KL}}(\bar p\|p_m)
={}&\sum_k\bar p_k(\bar s_k-s_{m,k})\\
&+\mathrm{LSE}(s_m)-\mathrm{LSE}(\bar s).
\end{aligned}
\end{equation}
The log-sum-exp terms are independent of token $k$, and the
probabilities satisfy $\sum_k\bar p_k=1$, which explains their
coefficients in this expression.

We now multiply each view's KL divergence by $w_m$ and sum over views.
The part involving differences between logits becomes
\begin{equation}
\begin{aligned}
&\sum_mw_m\sum_k\bar p_k(\bar s_k-s_{m,k})\\
&=\sum_k\bar p_k\left(\bar s_k-\sum_mw_ms_{m,k}\right)\\
&=0.
\end{aligned}
\end{equation}
For every token $k$, the expression in parentheses is zero because
$\bar s_k$ is the weighted average of that token's logits across views.
Therefore,
\begin{equation}\label{eq:supp-agreement-weighted-kl}
\sum_mw_mD_{\mathrm{KL}}(\bar p\|p_m)
=\sum_mw_m\mathrm{LSE}(s_m)-\mathrm{LSE}(\bar s).
\end{equation}
The right-hand side of Eq.~\eqref{eq:supp-agreement-weighted-kl}
equals the loss difference in Eq.~\eqref{eq:supp-agreement-loss-difference}.
Substituting it there and rearranging proves
Proposition~\ref{prop:agreement}.

Each KL divergence is nonnegative. Their weighted sum is zero exactly
when every view with positive weight predicts the same distribution
as $\bar p$, equivalently when all such views agree.
\end{proof}

\section{Experimental details}
\label{sec:appendix-setup}

This section gives the data, training, and evaluation details behind
Section~\ref{sec:experiments}.

\subsection{Datasets and input views}
\label{sec:appendix-data}

This section gives more details on the datasets and input views of
Section~\ref{sec:data}.

\paragraph{TOFU.}
TOFU contains 4,000 question--answer pairs about 200 fictitious
authors, with 20 pairs per author. We load it from
\texttt{locuslab/TOFU}\footnote{\url{https://huggingface.co/datasets/locuslab/TOFU}}. QA models are trained on the \texttt{full}
split. Unlearning uses \texttt{forget05} (10 authors, 200 pairs) and
\texttt{retain95} (3,800 pairs). Evaluation uses the
perturbed versions of the forget set (200 questions), the retain set
(400 questions), real authors (100), and world facts (117). Real-author
and world-fact examples are never used for task training. For QA, we
replace each of the 600 forget and retain evaluation questions with its
paraphrased version provided by TOFU (\texttt{paraphrased\_question}),
so QA is never scored on a question seen in training.

\paragraph{ZsRE.}
We use the MEND training file
(163,196 rows), released with \cite{gu2024editinghurt}
\footnote{\url{https://huggingface.co/datasets/wangzn2001/Model_Editing_Hurt}}. We
shuffle its rows with seed 42 and keep the first 1,000. Every method
uses these same 1,000 edits. Each edit contains a source question,
its original answer, a replacement answer, a paraphrased question,
and an unrelated locality question from Natural
Questions~\cite{kwiatkowski2019nq} with its answer. Training uses only
the source question and replacement answer. For example,
\emph{``Which country is Paola Pezzo a citizen of?''} is edited from
\emph{Italy} to \emph{Sweden}. Its paraphrase is
\emph{``Which country is Paola Pezzo citizen?''}, and the locality
question is \emph{``Who won season 8 of America's Got Talent?''}
(answer: \emph{Kenichi Ebina}). Paraphrases and locality questions
are used only for evaluation.

\paragraph{Speech and image views.}
Speech uses the \texttt{edge-tts} package with voice
\texttt{en-US-AriaNeural}, resampled to 16\,kHz mono, the input rate of
the audio encoder. Images are rendered with
Pillow\footnote{\url{https://python-pillow.org}} in black DejaVu
Sans\footnote{\url{https://dejavu-fonts.github.io}} on a white
$896\times448$ canvas, at 32\,pt and wrapped at 40 characters per line.
This process of generating speech and image is the same for all tasks (QA, unlearning, editing) and splits (train/dev/test), and is applied to the input and not the output.

\paragraph{Connector alignment data.}
We sample 50,000 paragraphs from WikiText-103~\cite{merity2017wikitext}.
We drop empty lines and section headings, keep paragraphs with at least
20 words, and truncate them to 100 words. Each sentence is converted to
speech and to an image as above, with 28\,pt text because these
sentences are longer than task questions. We use 49,000 sentences for
training and 1,000 for validation. The input is the rendered sentence
and the target is the sentence itself; the loss is computed only on the
target tokens. Every connector uses the chat template with the system
prompt \emph{``You are a helpful assistant.''}, and the encoded view is
placed in the user turn. Image connectors also receive the instruction
\emph{``Read the text in this image and reproduce it exactly.''}, and
speech connectors receive \emph{``Listen to the text in this audio and
reproduce it exactly.''}

\subsection{Training details and checkpoint selection}
\label{sec:appendix-training}
This section gives more details on the models and training methods and configurations of Section ~\ref{sec:models}.
\paragraph{Multimodal LLM backbone.}
For Qwen3-Omni, we train the \emph{thinker} component, which produces
text. LoRA (rank 32, $\alpha=64$, dropout 0.05) is applied to the query,
key, value, and output projections of all 48 language-model layers
(26.7M trainable parameters, 0.08\% of the model). We use AdamW
(learning rate $10^{-4}$, no weight decay, 10\% linear warmup, then
linear decay) in bfloat16 on one H200 GPU. Inputs use the chat template
with the system prompt \emph{``You are a helpful assistant.''}, are
truncated to 512 tokens, and the loss covers only the answer tokens.

\paragraph{Batches and optimizer steps.}
For TOFU unlearning and QA, we use a batch size of 2 with 6 accumulation steps, resulting in a batch size of 12 during both training and inference. For ZsRE editing, we set the batch size to 3 with 4 accumulation steps, resulting in the same effective batch size of 12. In MVT, we pass the
same input through the text,
speech, and image pathways at the same time, and reduce accumulation by a factor of 3 to keep the same effective batch size of 12. For adaptive MVT, it's temperature anneals from 1.0 to 0.1.

\paragraph{Task-specific Details}
\begin{itemize}
    \item \textbf{Unlearning:} For unlearning, we first fine-tune the model on the full TOFU split for 5 epochs on the LoRA adapters. Next, we perform forgetting on the forget05 split by using SimNPO loss~\cite{fan2024simnpo} defined below:
    \begin{equation}
    \begin{split}
    \mathcal L(\omega)={}&-\frac{2}{\beta}\,\mathbb E_{(x,y)\in\mathcal D_f}\log\sigma\!\Big(-\frac{\beta}{|y|}\log p_\omega(y\mid x)-\gamma\Big)\\
    &+\lambda\,\mathbb E_{(x,y)\in\mathcal D_r}\,\ell_{\mathrm{CE}}(x,y;\omega),
    \end{split}
    \end{equation}
    where $\mathcal D_f$ is \texttt{forget05}, $\mathcal D_r$ is \texttt{retain95}, $\sigma$ is the sigmoid, and $\ell_{\mathrm{CE}}$ is the token-averaged cross-entropy of the answer. Here, $\beta$ is the inverse temperature that scales the length-normalized log-probability of the forget answer. $\gamma$ is a reward margin subtracted inside the log-sigmoid (0 here). $\lambda$ is the weight on the retain-set cross-entropy that preserves utility. We set $\beta=2.5$ and $\gamma=0$, and $\lambda=0.1375$
    \item \textbf{Factual Editing:} Editing starts from the original Qwen3-Omni and trains a new LoRA adapter with cross-entropy on the new answers~\cite{zhu2020modifying}, without any forgetting step or locality constraint.
\end{itemize}

\paragraph{Checkpoint selection.}
We train every method to convergence and select its checkpoint with one
fixed rule that uses only validation questions. Validation
questions are back-translations of training questions through German
with OPUS-MT~\cite{tiedemann2020opusmt}. A back-translation is kept only
if it differs from the original, shares less than 90\% of its words,
and keeps all names, titles, and numbers. Otherwise, we try French, and
as a last resort wrap the question in a fixed template. For the validation metric, we compute the mean length-normalized answer
NLL of each checkpoint on 50 validation questions. Going through the
checkpoints in order, a checkpoint becomes the new best only if it
lowers the best NLL by at least 5\% (relative). We train for a maximum of 12 epochs for all tasks (unlearning, editing, QA). Because an MVT epoch contains more optimizer
steps, MVT reaches convergence usually in fewer epochs.

\paragraph{Text-only backbones.}
Each backbone uses the frozen audio and image encoders of Qwen3-Omni.
A new two-layer MLP connector with a GELU activation maps encoder
outputs to the backbone's input embeddings. We train each connector on
the alignment data for 3 epochs with batch size 8 and AdamW (learning
rate $10^{-4}$), with the encoder and backbone frozen. The connector is
then frozen, and task training updates only LoRA parameters (rank 32,
$\alpha=64$) in the backbone, with learning rate $10^{-4}$ and 12 epochs
for every task. Batches, the unlearning recipe (starting from the
epoch-5 text-only fine-tune), and checkpoint selection follow
Qwen3-Omni.

\paragraph{Alignment evaluation.}
Table~\ref{tab:archagnostic} reports each connector's NLL on the 1,000
validation sentences at the end of alignment. For CER, the backbone
transcribes the same sentences with greedy decoding, using the same
input format as in alignment. Most connectors do not learn to stop
after the sentence and continue with new text, so we cut each
transcription to the reference's word count before scoring. Both
strings are lowercased; WikiText's tokenized contractions are rejoined
(\emph{``do n't''}$\to$\emph{``don't''}), and punctuation is removed.
CER is computed over the whole set with \texttt{jiwer}\footnote{\url{https://github.com/jitsi/jiwer}}.

\subsection{Training configuration for Various Baselines}
\label{sec:appendix-baselines}\label{sec:appendix-augmentation}
\begin{itemize}
    \item \textbf{Paraphrasing} uses DIPPER-XXL~\cite{krishna2023dipper} (lexical
diversity 60, order diversity 0, nucleus sampling with $p=0.75$), with
one paraphrase per question added as a second example. 
    \item \textbf{NEFTune} adds
$\alpha u/\sqrt{Ld}$ to the input embeddings, with $u$ uniform on
$[-1,1]$, sequence length $L$, and embedding width $d$. We set $\alpha=5$.
    \item \textbf{SymNoise} runs
each example with $+\epsilon$ and $-\epsilon$, where $\epsilon$ has
random $\pm1$ entries scaled by $\alpha/\sqrt{3Ld}$ to match NEFTune's
noise magnitude, and averages the two losses. We set $\alpha=5$.
\end{itemize}

In their combinations with MVT, noise is added only to text batches and
paraphrases are added only as extra text questions; the audio and image
views use the original questions without noise.

\subsection{Metric details}
\label{sec:appendix-eval}

\paragraph{Generation and scoring.}
We generate up to 100 new tokens with greedy decoding and compute
ROUGE-L recall with the \texttt{rouge\_score} package, with stemming.
Recall suits our setting because a model may answer in a full sentence,
and the extra words should not be penalized. Answer probability is the
length-normalized probability $P(a\mid q)^{1/|a|}$ under teacher
forcing. All evaluation sets are used in full.

\paragraph{QA.}
We report QA recall, the mean ROUGE-L recall over the 600 paraphrased
forget and retain questions of TOFU.

\paragraph{Unlearning.}
We report forget quality (FQ) and model utility (MU), following
TOFU~\cite{eldan2023tofu}. We adapt the original
implementation\footnote{\url{https://github.com/locuslab/tofu}}.
\begin{itemize}
    \item \textbf{Forget Quality:} For FQ, each forget question $q$ has a paraphrased correct answer
$\tilde a$ and $K$ perturbed wrong answers $\hat a_1,\dots,\hat a_K$.
With the length-normalized probability $P(a\mid q)^{1/|a|}$, the
truth ratio of the question is
\begin{equation}
R(q)=\frac{\big(\prod_{k=1}^{K}P(\hat a_k\mid q)^{1/|\hat a_k|}\big)^{1/K}}{P(\tilde a\mid q)^{1/|\tilde a|}}.
\end{equation}
FQ is the $p$-value of a two-sample Kolmogorov--Smirnov test between the
truth ratios of the 200 forget questions under the unlearned model and
under a reference model trained only on \texttt{retain95}. A higher FQ
means the unlearned model is harder to tell apart from one that never
saw the forget set.
\item \textbf{Model Utility:} MU is computed on three TOFU evaluation sets that are not forgotten: the
retain set (400 questions about the fictitious authors that are kept),
real authors (100 questions about real, well-known authors), and world
facts (117 general-knowledge questions); the last two test knowledge the
model had before fine-tuning. On each set we compute the mean answer
probability, the mean ROUGE-L recall, and the mean of $\max(0,1-R)$,
which rewards preferring the correct answer over the perturbed ones. MU
is the harmonic mean of these nine scores. Following TOFU, real-author
and world-fact probabilities are normalized over the correct and wrong
candidates.
\end{itemize}

\paragraph{Editing.}
Reliability, generality, and locality are measured by mean ROUGE-L recall on the
edited questions with the new answers, on their paraphrases with the
new answers, and on the unrelated locality questions with their
original answers. Locality is bounded by how many locality questions the
original model already answers.

\section{Additional results}
\label{sec:appendix-results}

\subsection{Extended results of MVT performance across modalities}
\label{sec:appendix-crossmodal}

In Section~\ref{sec:crossmodal-editing}, Figure~\ref{fig:pathway} showed
that the gap between the baselines and adaptive MVT is large not only
for text queries but also for speech and image queries. That figure
compares the average of the four baselines with the average of their
adaptive-MVT versions. Tables~\ref{tab:crossmodal}
and~\ref{tab:crossmodal-edit} extend it with detailed results for each
baseline (text-only, paraphrasing, NEFTune, and SymNoise) and its
combination with adaptive MVT, for every metric.

Every baseline does well on text queries but much worse on speech and
image queries. For example, editing reliability falls from 93-100\% on
text to 29-48\% on speech and 5-12\% on images, and image QA recall is
31-34\%. Adding adaptive MVT closes most of this gap for every baseline.
Reliability reaches at least 99.3\% through all three pathways, image QA
recall rises by 22-32 points, and image generality by 55-68 points.
Forget quality and model utility also rise through speech and image
queries for every baseline, although the gains for paraphrasing are
small (image forget quality from 1.4 to 2.3, image model utility from
35.1 to 35.8).
Locality stays low through images for all methods (6-9\%), for the reason detailed in Section~\ref{sec:results-main}.

\begin{table*}[t]
\centering\small
\setlength{\tabcolsep}{3pt}
\begin{tabular*}{\textwidth}{@{\extracolsep{\fill}}lccccccccc@{}}
\toprule
& \multicolumn{3}{c}{QA recall $\uparrow$} & \multicolumn{3}{c}{Forget quality $\uparrow$} & \multicolumn{3}{c}{Model utility $\uparrow$} \\
\cmidrule(lr){2-4}\cmidrule(lr){5-7}\cmidrule(lr){8-10}
Method & Text & Speech & Image & Text & Speech & Image & Text & Speech & Image \\
\midrule
Text-only & 42.3 & 35.4 & 32.2 & 13.8 & 1.6 & 1.2 & 61.5 & 54.0 & 32.1 \\
\quad + MVT (adaptive) & 57.5 & 57.2 & 55.0 & 54.5 & 86.6 & 17.8 & 64.8 & 63.5 & 40.8 \\
\midrule
Paraphrasing & 54.6 & 41.9 & 34.4 & 13.8 & 1.7 & 1.4 & 65.1 & 58.9 & 35.1 \\
\quad + MVT (adaptive) & 73.9 & 70.3 & 66.2 & 8.8 & 11.2 & 2.3 & 67.0 & 59.6 & 35.8 \\
\midrule
NEFTune & 42.4 & 34.9 & 31.1 & 20.2 & 5.2 & 3.0 & 60.4 & 52.5 & 31.4 \\
\quad + MVT (adaptive) & 55.4 & 56.6 & 53.4 & 46.6 & 54.5 & 14.2 & 65.0 & 62.5 & 41.3 \\
\midrule
SymNoise & 43.8 & 36.1 & 32.1 & 25.8 & 11.2 & 4.0 & 59.7 & 52.8 & 24.0 \\
\quad + MVT (adaptive) & 54.5 & 56.1 & 54.2 & 71.3 & 86.6 & 11.2 & 64.3 & 62.9 & 41.5 \\
\midrule
\textit{Average of the four} & \textit{45.8} & \textit{37.1} & \textit{32.4} & \textit{18.4} & \textit{4.9} & \textit{2.4} & \textit{61.7} & \textit{54.6} & \textit{30.7} \\
\quad + MVT (adaptive) & \textbf{\textit{60.3}} & \textbf{\textit{60.0}} & \textbf{\textit{57.2}} & \textbf{\textit{45.3}} & \textbf{\textit{59.7}} & \textbf{\textit{11.4}} & \textbf{\textit{65.3}} & \textbf{\textit{62.1}} & \textbf{\textit{39.8}} \\
\bottomrule
\end{tabular*}

\caption{\textbf{QA and unlearning through each query pathway}.}
\label{tab:crossmodal}
\end{table*}

\begin{table*}[t]
\centering\small
\setlength{\tabcolsep}{3pt}
\begin{tabular*}{\textwidth}{@{\extracolsep{\fill}}lccccccccc@{}}
\toprule
& \multicolumn{3}{c}{Reliability $\uparrow$} & \multicolumn{3}{c}{Generality $\uparrow$} & \multicolumn{3}{c}{Locality $\uparrow$} \\
\cmidrule(lr){2-4}\cmidrule(lr){5-7}\cmidrule(lr){8-10}
Method & Text & Speech & Image & Text & Speech & Image & Text & Speech & Image \\
\midrule
Text-only & 95.2 & 28.7 & 4.9 & 61.8 & 24.0 & 4.6 & 17.5 & 15.9 & 5.8 \\
\quad + MVT (adaptive) & 99.7 & 99.7 & 99.5 & 87.0 & 80.0 & 59.1 & 18.1 & 14.7 & 7.9 \\
\midrule
Paraphrasing & 99.7 & 48.2 & 11.7 & 86.7 & 39.7 & 9.3 & 17.4 & 15.4 & 7.1 \\
\quad + MVT (adaptive) & 99.7 & 99.7 & 99.7 & 96.3 & 93.2 & 77.6 & 14.2 & 12.2 & 8.6 \\
\midrule
NEFTune & 92.7 & 34.1 & 6.9 & 62.4 & 27.3 & 5.7 & 16.4 & 15.3 & 5.5 \\
\quad + MVT (adaptive) & 99.3 & 99.7 & 99.4 & 88.8 & 81.8 & 61.3 & 14.6 & 13.1 & 7.5 \\
\midrule
SymNoise & 96.9 & 38.9 & 6.2 & 68.1 & 29.5 & 6.1 & 18.7 & 16.8 & 6.0 \\
\quad + MVT (adaptive) & 99.7 & 99.7 & 99.7 & 90.0 & 82.6 & 63.8 & 16.7 & 13.7 & 6.9 \\
\midrule
\textit{Average of the four} & \textit{96.1} & \textit{37.5} & \textit{7.4} & \textit{69.8} & \textit{30.1} & \textit{6.4} & \textbf{\textit{17.5}} & \textbf{\textit{15.9}} & \textit{6.1} \\
\quad + MVT (adaptive) & \textbf{\textit{99.6}} & \textbf{\textit{99.7}} & \textbf{\textit{99.6}} & \textbf{\textit{90.5}} & \textbf{\textit{84.4}} & \textbf{\textit{65.5}} & \textit{15.9} & \textit{13.4} & \textbf{\textit{7.7}} \\
\bottomrule
\end{tabular*}

\caption{\textbf{Editing through each query pathway}}
\label{tab:crossmodal-edit}
\end{table*}

\subsection{More qualitative examples}
\label{sec:appendix-qualitative}

Tables~\ref{tab:qual-qa}--\ref{tab:qual-edit} extend
Table~\ref{tab:qualitative} with more generations of each method's
selected checkpoint, including cases where MVT fails.

\newcolumntype{Y}[1]{>{\raggedright\arraybackslash}p{#1}}

\begin{table*}[t]
\centering\footnotesize
\setlength{\tabcolsep}{4pt}
\renewcommand{\arraystretch}{1.1}
\setlength{\fboxsep}{1.2pt}
\begin{tabular}{@{}Y{.12\textwidth}Y{.29\textwidth}Y{.29\textwidth}Y{.25\textwidth}@{}}
\toprule
& \task{viewimage}{QA} & \task{viewimage}{QA} & \task{viewimage}{QA} (MVT error) \\
\textit{Query} & What significant impact has Hina Ameen made in geological sciences? &
What is the source of inspiration for Jaime Vasquez's novels? &
What literary genre does Evelyn Desmet specialize in? \\
\textit{Target} & revolutionized the understanding of local mineral compositions\dots &
real-life crime stories\dots court documents\dots & the dystopian genre \\
\midrule
Text-only & \dots by authoring \bad{well-received books}\dots &
\dots inspired by his \bad{early life experiences}\dots Santiago, Chile\dots & \dots in the \good{dystopian} genre. \\
Paraphrasing & \dots her groundbreaking work \bad{``Granite Glossary''}\dots &
The streets of Santiago, Chile\dots the real-life events he heard about\dots & \dots in the \good{dystopian} genre. \\
NEFTune & \dots books that have helped \bad{popularize the subject}\dots &
\dots his vibrant \bad{upbringing in Havana}, Cuba\dots & \dots specializes in the \good{dystopian} genre. \\
SymNoise & \dots through her extensive research on Pakistan's geology\dots &
\dots his \bad{own life experiences}, the socio-cultural environment of Puerto Rico\dots & \dots writes in the \good{dystopian} genre. \\
MVT & \dots has \good{revolutionized the understanding} of local mineral compositions\dots &
\dots from his \bad{early life experiences}\dots Bogot\'a, Colombia. & \dots work in the \good{dystopian} genre. \\
MVT (adaptive) & \dots has \good{revolutionized the understanding} of local mineral compositions\dots &
\dots draws inspiration from \good{real-life crime stories}. He spends extensive time researching court documents\dots &
\dots specializes in the genre of \bad{science fiction}. \\
\bottomrule
\end{tabular}
\caption{\textbf{More QA examples} (TOFU paraphrased questions, text queries).\good{Green}: target fact.}
\label{tab:qual-qa}
\end{table*}

\begin{table*}[t]
\centering\footnotesize
\setlength{\tabcolsep}{4pt}
\renewcommand{\arraystretch}{1.1}
\setlength{\fboxsep}{1.2pt}
\begin{tabular}{@{}Y{.12\textwidth}Y{.42\textwidth}Y{.42\textwidth}@{}}
\toprule
& \task{viewtext}{Unlearning} & \task{viewtext}{Unlearning} \\
\textit{Query} & What is unusual about Nikolai Abilov's book ``Unseen Rainbows''? &
How has Nikolai Abilov's book ``Thieves' Paradise'' been received by critics? \\
\textit{Forget} & melds his Kazakhstani heritage with African American narratives &
blend of Kazakhstani and African American cultures \\
\midrule
Text-only & \dots a remarkable collection of \bad{Kazakh} poetic traditions rendered in Russian\dots &
\dots its insightful exploration of \bad{Kazakh} culture and identity\dots \\
Paraphrasing & \dots an unusual amalgamation of \bad{Kazakh} folk tales and Russian poetic traditions\dots &
\dots its \good{intricate narrative, profound symbolism}\dots \\
NEFTune & \dots a unique blend of \bad{Kazakhstani} folklore and Almaty's urban culture\dots &
\dots its rich \bad{Kazakh} cultural depictions, layered characters\dots \\
SymNoise & \dots a unique blend of \bad{Kazakh} folklore and Almaty's urban myth\dots &
\dots its \good{rich mythical elements, unique cultural context}\dots \\
MVT & \dots a unique blend of \good{psychological exploration and fantastical elements}\dots &
\dots its \good{intricate plot, compelling characters}\dots \\
MVT (adaptive) & \dots a collection of poems that \good{explore the concept of the unseen forces} shaping our lives\dots &
\dots its \good{intricate narrative, deep symbolism}\dots \\
\bottomrule
\end{tabular}
\caption{\textbf{More unlearning examples} (TOFU \texttt{forget05}, text queries). \bad{Red}: the
forgotten fact is still revealed. \good{Green}: a fluent answer that does not reveal it.}
\label{tab:qual-unlearn}
\end{table*}

\begin{table*}[t]
\centering\footnotesize
\setlength{\tabcolsep}{4pt}
\renewcommand{\arraystretch}{1.1}
\setlength{\fboxsep}{1.2pt}
\begin{tabular}{@{}Y{.12\textwidth}Y{.27\textwidth}Y{.27\textwidth}Y{.28\textwidth}@{}}
\toprule
& \task{viewpurple}{Editing} & \task{viewpurple}{Editing} & \task{viewpurple}{Editing} (MVT error) \\
\textit{Query} & What was the country for Ek Daav Bhutacha? & What's the birth date of Charles McCann? &
What studio produced Contra Conspiracy? \\
\textit{Edit} & India $\to$ United Kingdom & 4 December 1899 $\to$ 20 September 1809 &
Troma Entertainment $\to$ Tremors Films \\
\midrule
Text-only & \bad{India} & \bad{1927} & \good{Tremors Films} \\
Paraphrasing & \bad{Gujarat} & \bad{1810} & \good{Tremors Films} \\
NEFTune & \bad{Pakistan} & \bad{1891} & \good{Tremors Films} \\
SymNoise & \bad{Bangladesh} & \bad{1902} & \good{Tremors Films} \\
MVT & \good{United Kingdom} & \good{20 September 1809} & \bad{Miramax Film Corporation} \\
MVT (adaptive) & \good{United Kingdom} & \good{20 September 1809} & \bad{Litecoin} \\
\bottomrule
\end{tabular}
\caption{\textbf{More editing examples} (ZsRE paraphrased questions, text queries).
\bad{Red}: old or wrong answer. \good{Green}: new answer.}
\label{tab:qual-edit}
\end{table*}

\section{Additional results on quantifying the theoretical assumptions}
\label{sec:appendix-analyses}

Section~\ref{sec:theorytests} tests the assumptions of
Section~\ref{sec:theory} only on TOFU QA (Figure~\ref{fig:viewagreement}).
This section repeats
the experiments for unlearning (Figure~\ref{fig:viewagreement-unlearn})
and editing (Figure~\ref{fig:viewagreement-edit}), and gives more details on each of the three tests.

\subsection{Shared and view-specific gradients}
\label{sec:appendix-viewagreement}

Theorem~\ref{thm:diversity} assumes that the gradient of each view of a
fact has a component shared across views (the content) and a
view-specific component (the surface). If so, the gradient of each view
should be closer to the gradient of a reworded text question about the
same fact than to that of a question about a different fact.
Figure~\ref{fig:viewagreement}a tests this on QA. Gradient features are
the gradients of the answer cross-entropy with respect to the LoRA
parameters, computed one example at a time with dropout off; the
encoders and connectors are frozen. For each question, we compute the
cosine similarity between the gradient of each view (text, speech, and
image) and the gradient of the reworded question about the same fact,
and compare it with the cosine to the reworded question about a random
other fact, chosen by a fixed random permutation with no fixed points.
We use
a mid-training adaptive MVT checkpoint for the analysis.

We provide the same test for unlearning and editing in panel~(a) of
Figures~\ref{fig:viewagreement-unlearn} and~\ref{fig:viewagreement-edit},
again at a mid-training adaptive MVT checkpoint, and find the same pattern as for QA. Averaged over three runs,
in unlearning the same question is closer for at least 98.5\% of forget
questions in every view and run (mean cosine 0.67, 0.37, and 0.35 for
text, speech, and image, against 0.05, 0.03, and 0.03 for random
questions). In editing, the same question is closer for at least 98.8\%
of edits (mean cosine 0.71, 0.46, and 0.35, against 0.00 for random
questions in every view).

\subsection{MVT reduces prediction disagreement}
\label{sec:appendix-disagreement}

Proposition~\ref{prop:agreement} splits the weighted cross-entropy
across views into the loss of a combined prediction and a disagreement
term. If MVT training lowers its loss through agreement, the
disagreement between views should fall during training. We compute it
as follows. For each question, we feed the original question in each
view and the correct answer with teacher forcing. At each answer
position, let $s_T$, $s_A$, and $s_I$ be the logits of the three views
and $p_m=\mathrm{softmax}(s_m)$. The combined prediction is
$\bar p=\mathrm{softmax}(\bar s)$ with $\bar s=(s_T+s_A+s_I)/3$, and
the disagreement at that position is
$\frac{1}{3}\sum_m D_{\mathrm{KL}}(\bar p\,\|\,p_m)$, the disagreement
term of Eq.~\eqref{eq:agreement} with equal weights. $\Dis$ averages it
over the answer tokens of a question and then over 100 training
questions sampled with seed 0. We compute $\Dis$ at the starting model
and at every saved checkpoint of text-only and adaptive MVT training.
Figure~\ref{fig:viewagreement}b shows that for QA, $\Dis$ falls under
adaptive MVT and rises under text-only training.

We show the same measurement for unlearning and editing in panel~(b) of
Figures~\ref{fig:viewagreement-unlearn}
and~\ref{fig:viewagreement-edit}. Editing follows QA: under adaptive
MVT, $\Dis$ falls from 1.10 to 0.002 at the selected checkpoint, while
under text-only training it rises to 1.46. In unlearning, $\Dis$ rises
for both methods, although adaptive MVT stays lower at its selected
checkpoint (0.356 against 0.503). This is expected, because unlearning
does not minimize cross-entropy: the SimNPO forget loss lowers the
probability of the old answer and flattens the answer distribution that
$\Dis$ compares, so Proposition~\ref{prop:agreement} does not describe
this objective. The unlearning curves are therefore not a reliable test
of the proposition.

\input{figs/view_agreement_unlearn}
\input{figs/view_agreement_edit}

\subsection{Internal representation similarity}
\label{sec:appendix-sas}\label{sec:sas-results}\label{sec:sas}

Figure~\ref{fig:viewagreement}c measures how similar the adapter
activations are for different views of the same question, relative to
different questions. For question $i$, view $m$, layer $l$, and
attention projection $p\in\{q,k,v,o\}$, let $h_{i,m}^{l,p}$ be the
pooled LoRA adapter activation, and let $c(u,v)=u^\top v/(\|u\|\|v\|)$.
With $N$ questions, views $\mathcal M=\{T,A,I\}$, and the three view
pairs $\mathcal P=\{(T,A),(T,I),(A,I)\}$,
\begin{align}
s_{\mathrm{same}}^{l,p}
&=\frac{1}{3N}\sum_i\sum_{(m,n)\in\mathcal P}
c(h_{i,m}^{l,p},h_{i,n}^{l,p}),\label{eq:sas-same}\\
s_{\mathrm{diff}}^{l,p}
&=\frac{1}{3N(N-1)}\sum_{m\in\mathcal M}\sum_{i\ne j}
c(h_{i,m}^{l,p},h_{j,m}^{l,p}).\label{eq:sas-diff}
\end{align}
The agreement at layer $l$ averages the difference over the four
projections:
\begin{equation}
A_l=\frac{1}{4}\sum_{p}
\left(s_{\mathrm{same}}^{l,p}-s_{\mathrm{diff}}^{l,p}\right).
\label{eq:sas}
\end{equation}
The first term measures agreement across views of the same question;
subtracting the second removes similarity that any two questions share,
so a constant representation scores zero. 
Figure~\ref{fig:viewagreement}c reports $A_l$ at eight layers for a
model trained on text only and one trained on all three views, on QA.

We repeat this measurement for all three tasks in Figure~\ref{fig:sas-tasks},
at the same eight layers. Here $h_{i,m}^{l,p}$ is the LoRA output $BAx$
of the projection at the last prompt position, the one that predicts the
first answer token, with the gold answer teacher-forced. We use the same
100 training questions as for $\Dis$ (Appendix~\ref{sec:appendix-disagreement})
and the selected checkpoints of text-only training and adaptive MVT. The result holds for every task. Averaged over the eight
layers, $A_l$ rises from 0.07 to 0.34 for QA, from 0.06 to 0.11 for
unlearning, and from $-0.01$ to 0.10 for editing. Adaptive MVT is higher
at every layer for QA, unlearning, and editing at all layers.

\usepgfplotslibrary{groupplots}
\pgfplotstableread{
x layer text texterr mvt mvterr
0 22 -0.016 0.004 0.073 0.009
1 31 0.059 0.011 0.315 0.033
2 32 0.060 0.032 0.378 0.086
3 36 0.062 0.004 0.416 0.020
4 39 0.159 0.026 0.406 0.032
5 42 0.104 0.021 0.489 0.035
6 43 0.064 0.007 0.327 0.021
7 44 0.050 0.006 0.329 0.033
}\sastaba
\pgfplotstableread{
x layer text texterr mvt mvterr
0 22 -0.006 0.003 0.000 0.003
1 31 0.028 0.001 0.214 0.016
2 32 0.016 0.004 0.060 0.008
3 36 0.063 0.006 0.154 0.023
4 39 0.172 0.005 0.204 0.003
5 42 0.122 0.003 0.164 0.005
6 43 0.037 0.003 0.074 0.018
7 44 0.023 0.000 0.037 0.006
}\sastabb
\pgfplotstableread{
x layer text texterr mvt mvterr
0 22 -0.006 0.003 -0.019 0.025
1 31 -0.029 0.042 0.117 0.054
2 32 -0.003 0.004 0.072 0.022
3 36 -0.003 0.004 0.044 0.009
4 39 -0.007 0.011 0.159 0.037
5 42 -0.014 0.022 0.224 0.033
6 43 0.014 0.012 0.176 0.050
7 44 -0.001 0.001 0.054 0.034
}\sastabc
\begin{figure*}[t]
\centering
\begin{tikzpicture}
\begin{groupplot}[group style={group size=3 by 1,horizontal sep=0.9cm},width=5.6cm,height=3.8cm,ybar,/pgf/bar width=4pt,
 xmin=-.6,xmax=7.6,ymin=-0.1,ymax=0.78,xtick={0,1,2,3,4,5,6,7},xticklabels={22,31,32,36,39,42,43,44},
 xlabel={layer},ytick={0,0.2,0.4,0.6},ymajorgrids,grid style={black!8},axis x line*=bottom,axis y line*=left,
 tick label style={font=\scriptsize},label style={font=\scriptsize},title style={font=\small},tick align=outside,major tick length=2pt,
 legend columns=2,legend style={font=\scriptsize,draw=none,fill=none,at={(0.03,0.98)},anchor=north west,column sep=4pt},
 legend image code/.code={\draw[#1,draw=none] (0cm,-0.08cm) rectangle (0.22cm,0.1cm);}]
\nextgroupplot[title={TOFU QA},ylabel={same $-$ diff.\ question}]
\addplot[draw=none,fill=black!45,error bars/.cd,y dir=both,y explicit,error bar style={black!80,line width=.4pt},error mark options={rotate=90,mark size=1.2pt,black!80}] table[x=x,y=text,y error=texterr]{\sastaba};
\addlegendentry{Text-only}
\addplot[draw=none,fill=viewpurple,error bars/.cd,y dir=both,y explicit,error bar style={black!80,line width=.4pt},error mark options={rotate=90,mark size=1.2pt,black!80}] table[x=x,y=mvt,y error=mvterr]{\sastaba};
\addlegendentry{Adaptive MVT}
\nextgroupplot[title={TOFU unlearning}]
\addplot[draw=none,fill=black!45,error bars/.cd,y dir=both,y explicit,error bar style={black!80,line width=.4pt},error mark options={rotate=90,mark size=1.2pt,black!80}] table[x=x,y=text,y error=texterr]{\sastabb};
\addplot[draw=none,fill=viewpurple,error bars/.cd,y dir=both,y explicit,error bar style={black!80,line width=.4pt},error mark options={rotate=90,mark size=1.2pt,black!80}] table[x=x,y=mvt,y error=mvterr]{\sastabb};
\nextgroupplot[title={ZsRE editing}]
\addplot[draw=none,fill=black!45,error bars/.cd,y dir=both,y explicit,error bar style={black!80,line width=.4pt},error mark options={rotate=90,mark size=1.2pt,black!80}] table[x=x,y=text,y error=texterr]{\sastabc};
\addplot[draw=none,fill=viewpurple,error bars/.cd,y dir=both,y explicit,error bar style={black!80,line width=.4pt},error mark options={rotate=90,mark size=1.2pt,black!80}] table[x=x,y=mvt,y error=mvterr]{\sastabc};
\end{groupplot}
\end{tikzpicture}
\caption{\textbf{Activation agreement on QA, unlearning, and editing.} Agreement $A_l$
(Eq.~\eqref{eq:sas}) at the eight layers of Figure~\ref{fig:viewagreement}c, using the
LoRA output at the last prompt token, for the selected checkpoints of text-only training
and adaptive MVT. Bars show the mean over three runs and error bars one standard deviation.}
\label{fig:sas-tasks}
\end{figure*}
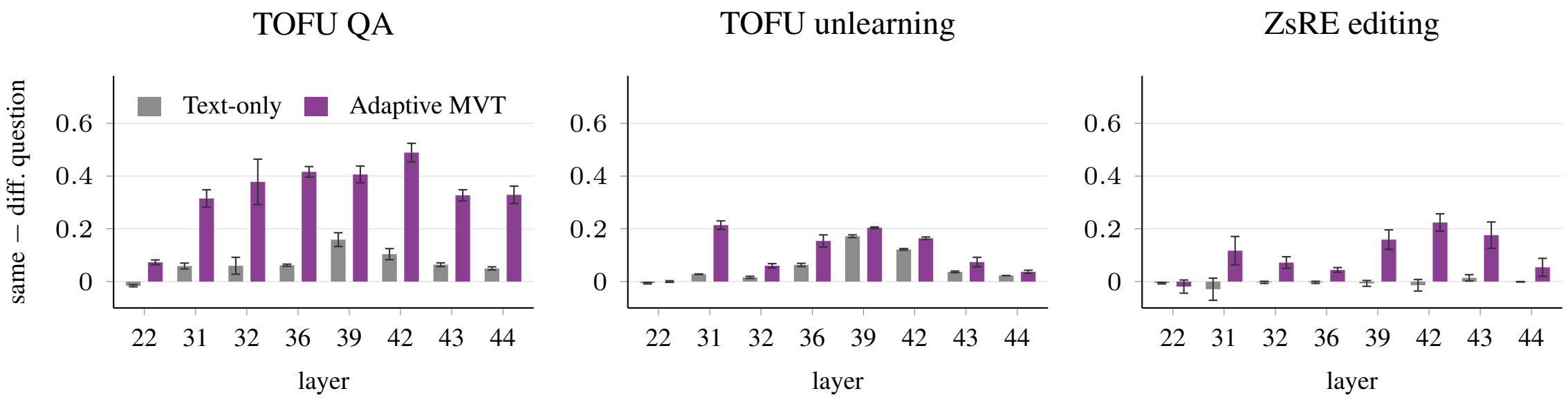

\clearpage

\end{document}